# Hybrid Magnetically and Electrically Powered Metallo-Dielectric Janus Microrobots: Enhanced Motion Control and Operation Beyond Planar Limits

Ido Rachbuch[1], Sinwook Park[1], Yuval Katz[1], Touvia Miloh[1], Gilad Yossifon[1,2]

[1]School of Mechanical Engineering, Tel-Aviv University, Israel

[2]Department of Biomedical Engineering, Tel-Aviv University, Israel.

Correspondence and requests for materials should be addressed to G.Y. (email: gyossifon@tauex.tau.ac.il)

**Abstract**

This study introduces the integration of hybrid magnetic and electric actuation mechanisms to achieve advanced motion capabilities for Janus particle (JP) microrobots. We demonstrate enhanced in-plane motion control through versatile control strategies and present the concepts of interplanar transitions and 2.5-dimensional (2.5D) trajectories, enabled by magnetic levitation and electrostatic trapping. These innovations expand the mobility of JPs into 3D space, allowing dynamic operation beyond the limitations of traditional surface-bound motion. Key functionalities include obstacle crossing, transitions to elevated surfaces, and discrete surface patterning enabling highly localized interventions. Using this set of tools, we also showcase the controlled out-of-plane transport of both synthetic and biological cargo. Together, these advancements lay the groundwork for novel microrobot-related applications in microfluidic systems and biomedical research.

**Introduction**

Hybrid micromotors, combining various propulsion mechanisms (e.g. electric and thermal[1], chemical[2], biological and magnetic[3], acoustic and optical[4]), have shown significant potential in various environmental and biomedical applications due to their enhanced operational robustness, complementary actuation control, and ease of integrating additional functionalities[5–7]. Among these, electrically powered micromotors offer distinct capabilities, such as selective cargo transport[8,9], sensing[7], and local stimulation[10], controllable by adjusting the applied alternating current (AC) frequency. Similarly, magnetically powered micromotors and microrobots have been extensively investigated for their ability to be remotely operated with high precision, under various forms of external magnetic fields[11], and across diverse environmental conditions, including varying chemical (e.g. pH, conductivity) solution properties[12], and resistive fluid flows[13,14]. Recognizing the advantages of these two modalities, a hybrid system that combines magnetic and electric powering could offer increased versatility and unique capabilities. Such a hybrid combination is especially valuable due to the quenching of electrokinetic propulsion with increasing solution conductivity[15], which practically ceases the propulsion at solution conductivities exceeding[16,17] ≈0.3 mS cm$^{-1}$. Magnetic propulsion can compensate for this limitation, thereby extending the applicability of the micromotor to a broader range of solution conductivities, as demonstrated in our recent study on the hybrid JP micromotor[18]. However, to maximize the potential of such a hybrid system, it is essential to understand the various possible combinations of the contributing components, each with its own benefits, and to integrate them effectively for enhanced control and optimized functional performance.

In this study, we focus on metallo/magneto-dielectric Janus particles (JPs) as one of the simplest models for a hybrid micromotor powered by magnetic and electric fields. These particles are widely studied in micromotor research due to their straightforward asymmetric structure, which allows for directional motion, as well as their ease of fabrication, making them an ideal platform for exploring hybrid actuation mechanisms[9,18–21]. By applying synergistic magnetic and electric field combinations, we aim to achieve enhanced, controlled motion of these active particles, utilizing to the fullest some of their unrealized potential. Previous studies have implemented magnetic field-based steering of electrically propelled particles[7,9,10,19], and fully magnetic propulsion and steering via rolling motion driven by rotating magnetic fields[13,22–24]. In the latter case, our recent study also examined the contribution of an added electric field component during magnetic rolling, demonstrating enhanced mobility as well as cargo manipulation and sensing capabilities[18]. Here, we employ these two operational strategies and also introduce a new navigation scheme: fixing the rotation axis of the magnetic field to induce linear rolling motion (thereby modeling simplified operating conditions using a permanent magnet with a single rotation axis), while adding an orthogonal velocity component through electric propulsion. We systematically characterize each of the proposed magnetic-electric control strategies by demonstrating open-loop control, discussing relevant parameters, and introducing closed-loop navigation based on tailored control algorithms. To achieve this, we developed a comprehensive control system leveraging software and hardware, enabling either manual manipulation or automated manipulation relying on real-time image analysis and trajectory planning. In this context, due to their ability to precisely navigate and perform programmable actions, JPs evolve from being merely active particles (or micromotors) to being more aptly regarded as 'microrobots'[25,26].

Until now, the motion of metallo-dielectric JPs has largely been confined to a two-dimensional (2D) plane – typically the floor of a fluid-filled microchamber or microfluidic chip[7,8,10,12,18]. This restriction arises not only from gravity pulling downward, but also from the dependency of common actuation mechanisms (e.g. magnetic rolling[12,13], self-dielectrophoresis[15] and other mechanisms of similar micromotors[27–29]) on interaction with a nearby physical boundary. Such interaction is categorized as 'surface-assisted' motion[11], with variations like 'surface-rolling'[13,21] and 'surface walking'[28].. Consequently, most prior studies maintained continuous surface contact, addressing three-dimensional (3D) locomotion by traversing inclined boundaries[30,31] or designing geometries capable of climbing specific topologies[14,29]. Here, we introduce a novel approach that adds another dimension of motion - through interplanar transitions. By leveraging magnetic field gradients and electrostatic trapping, we levitate JPs to the ceiling of a microchamber and hold them against gravity when the magnetic gradients are deactivated. This enables similar actuation mechanisms to operate on the ceiling, significantly expanding the JPs' motion capabilities. To the best of our knowledge, this is the first demonstration of 3D navigation based on combined magnetic and electric powering that doesn't require strong magnetization of the microrobot or sustained magnetic field gradients.

**Results**

***Magnetic and Electric Field Combinations for Enhanced Janus Particle (JP) Motion Control***

**Fig. 1A** illustrates some of the various combinations of electric and magnetic field actuations for controlling the motion of metallo/magneto-dielectric Janus Particles (JPs) both in-plane and out-of-plane, enabling them for example to cross barriers or descend onto elevated surfaces. Examples include electric propulsion combined with magnetic steering, and magnetic rolling coupled with electric alignment. A uniform AC electric field applied between parallel ITO-coated glass slides drives the in-plane self-propulsion of JPs. Depending on the applied frequency, JPs propel with either their dielectric side (Polystyrene) or metal-coated side (Ni-Au) facing forward, exhibiting induced charge electrophoretic (ICEP)[16] or self-dielectrophoretic (sDEP) propulsion[15], respectively. The magnetic field, generated by a commercially available electromagnet array system (MagnebotiX), provides enhanced versatility. It enables the application of uniform fixed or rotating fields for in-plane actuation, as well as magnetic field gradients to levitate JPs toward the ceiling, where they can be secured using electrostatic trapping.

Experiments were conducted with JPs measuring 10 and 27 μm in diameter within designated microchambers, with their motion controlled in both open-loop and closed-loop configurations. For open-loop control, we developed an integrated software and hardware system (see **Methods** section) to simultaneously manage the AC electric field and the magnetic field, allowing intuitive manual manipulation via keyboard inputs (**Fig. 1B**, **Supplementary Fig. 1**). Magnetic actuation parameters included field direction, magnitude, and rotation frequency (for magnetic rolling), while electric propulsion velocity and alignment depended on voltage and frequency, enabling versatile control of motion in various directions and magnitudes. By adjusting these parameters through our control system, we achieved precise in-plane microrobot navigation and utilized magnetic field gradients for interplanar transitions. For closed-loop control, we incorporated high level operations such as real-time tracking and path planning. Desired paths were defined as sequences of interconnected waypoints, which could be arbitrarily selected or derived from predefined shapes.

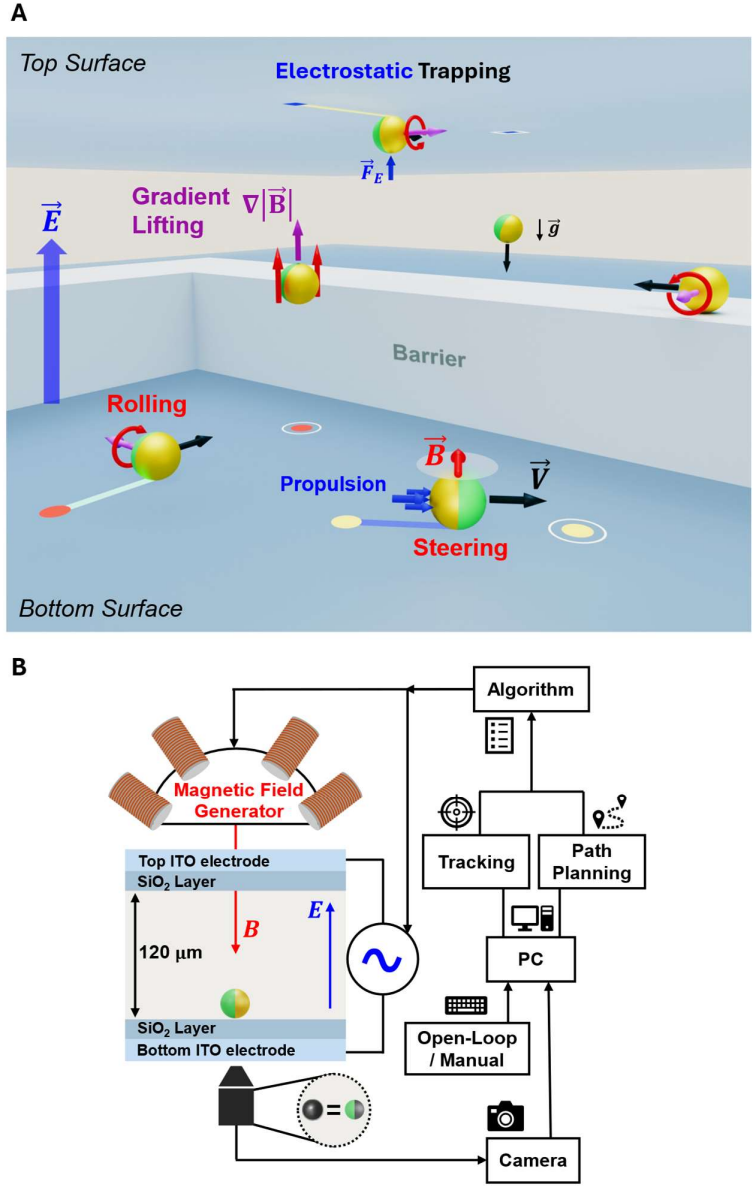

**Fig. 1. Hybrid magnetic and electric field actuations for enhanced motion control of metallo/magneto-dielectric Janus particle (JP) microrobots. A** Schematic illustration of various actuation methods in a 3D environment. A uniform AC electric field is applied between parallel ITO-coated glass slides, aligning the JPs along their dielectric-metal interface and inducing propulsion orthogonal to the field direction. A magnetic field, generated by an array of electromagnets, enables steering, rolling, and lifting of the microrobots through fixed, rotating, and gradient fields, respectively. Once lifted to the ceiling of the chamber, the microrobot is held against gravity by electrostatic attraction and transported horizontally using similar actuation mechanisms. This enables the microrobot to switch planes of motion between the bottom and the top substrates, overcome barriers, and descend onto elevated surfaces. **B** Experimental setup and control system diagram. Under optical microscopy the metal-coated hemisphere of the JP (Ni+Au) appears darker than the dielectric hemisphere. Real-time tracking and path-planning modules facilitate programmable navigation through a series of waypoints, depicted in (A) as circles and squares on the bottom and top surfaces, respectively.

Since the JPs were not pre-magnetized in a specific direction, their thin ferromagnetic Ni deposited layer exhibited dynamic magnetization induced solely by the applied external magnetic fields (up to 20 mT) during manipulation, with some residual magnetization likely persisting due to magnetic hysteresis[32]. Previous studies examined magnetization hysteresis loops of sputtered Ni films on similar Janus particles magnetized to saturation under strong magnetic fields (up to 1.8 T) and reported coercivity of 20 mT[13]. This suggests that the relatively low magnetic field magnitudes used here, well below saturation levels, were sufficient to dynamically alter magnetization of the initially non-magnetized particles. Under uniform (i.e. gradient-free) magnetic fields, the induced magnetization drove the particles into energetically favorable orientations, such that the long axis of the Ni-coated hemisphere, which lies approximately parallel to the JP's interface plane (see **Supplementary Fig. 2**), would tend to align with the direction of the magnetic field[33,34]. Similarly, under the uniform AC electric field applied between the parallel ITO electrodes, an electric dipole is induced in the Au layer of the JP, aligning the long axis of the metallic hemisphere with the direction of the field. This alignment is driven by electrostatic torque counteracting gravitational torque to establish a stable equilibrium[15,35]. The geometric symmetry of the JP around the axis normal to its metallo-dielectric interface plane (i.e. the symmetry axis) allows rotation about this axis without disturbing the alignment. As a result, both the magnetic and electric fields introduce a single rotational constraint, each reducing the particle's rotational degrees of freedom (DOFs) by one (see **Supplementary Fig. 3**). When both fields are applied simultaneously in non-parallel directions, their combined alignment constraints reduce the available three rotational DOFs to just one. Since the electric field is always parallel to the z-axis (normal to the ITO-coated substrates), the remaining two DOFs can be utilized by the magnetic field to induce rotations around either the z-axis or the JP's symmetry axis. Given that electric propulsion drives translation toward one of the JP's hemispheres, this configuration enables magnetic field-induced steering around the z-axis and rolling motion orthogonal to the electric field-driven propulsion. This arrangement minimizes interference and fosters a synergistic interaction between the two actuation mechanisms.

***Realization of Hybrid Magnetic and Electric Control Strategies for in-Plane Actuation***

**Fig. 2** demonstrate various strategies for combining magnetic and electric field actuations to achieve planar motion of JPs, showcasing the closed-loop implementation of each strategy with tailored algorithms and parameters. Each approach was tested in a controlled navigation session along a modified Bernoulli's lemniscate path with two sharp corners, used here as a standard benchmark[36–40] to qualitatively assess accuracy and highlight distinctive characteristics.

**Fig. 2A** compares the benchmark case of full magnetic rolling and steering, with the same setup under a high-frequency electric field, introducing several notable effects. The electric field acts on the induced electric dipole at the Au layer of the JPs and strongly aligns their Polystyrene-metal interface orthogonal to the rolling surface, thereby suppressing wobbling motion[18]. Additionally, the electric field enhances the effective friction coefficient with the substrate due to electrostatic attraction, resulting in increased linear motion for the same magnetic field rotation frequency[18,21]. While these effects generally improve rolling efficiency, the application of the electric field also leads to a decrease in the magnetic rolling step-out frequency, usually regarded as less favorable. The step-out frequency is reached when the magnetic torque is not strong enough to keep the microrobot's rotation synchronized with the magnetic field[41] . Without an electric field, this is mostly because of the resistance from the viscous force exerted by the surrounding fluid and the hydrodynamic interaction with the wall[42]. When the electric field is present, the observed decrease in the step-out frequency (especially with 27 μm JPs) indicates larger resistance to rotation, which could be attributed to stronger interactions with the substrate[17]. However, by carefully balancing the voltage, the magnetic field's magnitude, and its rotation frequency, it is possible to harness the beneficial effects of the combined actuation (e.g. increased rolling velocity, stabilized motion for effective cargo loading[18]), while avoiding the step-out region.

The proposed closed-loop control algorithm operates consistently in both fully magnetic and combined actuation scenarios. The algorithm is defined in a straightforward manner, utilizing the capability of the MagnebotiX system and similar electromagnet arrays to generate arbitrary magnetic fields rotating around any rotation axis. Since the rolling velocity vector normally aligns with the intersection of the rotation plane and the rolling surface, the required instantaneous rotation axis $\vec{R}_{ax}$ can be derived directly from the distance vector $\vec{d}$ connecting the microrobot's current position with the desired target, according to the following equation: $\vec{R}_{ax} = \hat{z} \times \vec{d}$, where the $z$-axis is perpendicular to the image plane. To add some level of velocity control, the magnetic field rotation frequency ($\omega$) was made proportional to the distance from the target waypoint, slowing the JP as it approached the target to avoid overshooting. Two selected values $\omega_1$ and $\omega_2$ defined a continuous linear range $[\omega_1, \omega_2]$ from which $\omega$ was sampled as a function of the instantaneous distance $|\vec{d}|$, such that $\omega = \omega_1 + \frac{|\vec{d}|}{d_{max}}(\omega_2 - \omega_1)$, where $d_{max}$ is the maximal possible distance in the trajectory. Selecting $\omega_1 = \omega_2$ would result in a constant rotation frequency. Comparing the performance of the same particle actuated by magnetic rolling with and without the electric field clearly shows that with the electric field the linear velocity was increased, leading to a shorter trajectory period by a factor of almost 2 (see trajectory images left of **Fig. 2A**) . The increased velocity resulted in minor overshoots over the sharp corners of the path, which could be avoided with slower rotation.

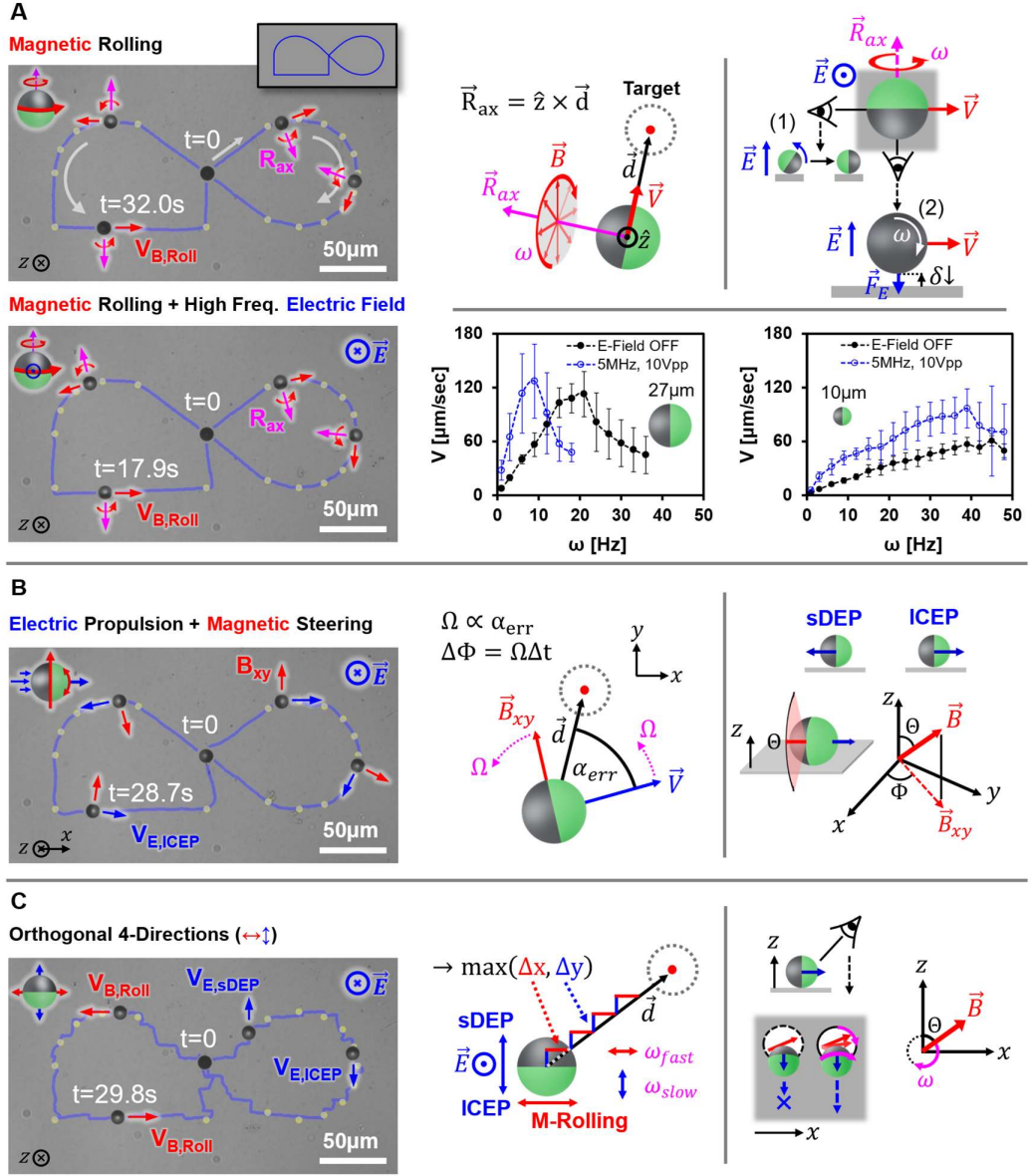

**Fig. 2. Hybrid magnetic and electric control strategies – parameters and closed-loop implementation.** Left - Closed-loop control sessions along a modified lemniscate path (blue) featuring two sharp corners, guided by a series of sampled waypoints (yellow dots). All sequences were conducted using the same 10 μm JP at 20 fps with a corresponding 20 Hz control rate, using a 3.5 mT magnetic field. Control parameters were selected to push performance limits and highlight differences between strategies. Right - Schematic representations of the control parameters and algorithms, emphasizing key effects and considerations. **A** Magnetic rolling (8-14 Hz) with or without high-frequency electric field (5 MHz, 10 $V_{pp}$), including systematic measurements of rolling velocity as a function of the magnetic field (3.5 mT) rotation frequency under both conditions. **B** Electric propulsion (ICEP at 2 kHz or sDEP at 50 kHz, both at 10 $V_{pp}$) combined with magnetic steering. sDEP. **C** Orthogonal 4-Directions – magnetic rolling ($\omega_{slow}$=0.5 Hz, $\omega_{fast}$=5 Hz) with a high-frequency electric field (5 MHz, 12 $V_{pp}$) restricted to the horizontal axis, while electric propulsion (ICEP at 2 kHz or sDEP at 50 kHz, both at 12 $V_{pp}$) generates perpendicular movement.

**Fig. 2B** illustrates electric propulsion governed by either the ICEP or sDEP mechanisms, depending on the AC field frequency, and guided by magnetic steering. The direction of motion is controlled by adjusting the angle of the planar magnetic field component $\vec{B}_{xy}$, which corresponds to the azimuthal angle Φ of the field $\vec{B}$ in spherical coordinates. This approach relies on the alignment of the JP's dielectric-metal interface plane relative to the x-axis around the z-axis, which is approximately orthogonal to the propulsion direction parallel to the substrate. The inclination angle of the magnetic field, Θ, plays a crucial role in propulsion. While a fully planar field (Θ=90°) might seem ideal for steering, our experiments revealed that in many cases propulsion efficiency varies significantly with Θ. Certain inclination angles optimize velocity, whereas others, including Θ=90°, can cause a complete halt. In each case, one particle propels efficiently while the other remains stationary due to differences in inclination angles.

Theoretically, given the rotational DOF of the JP's interface plane around the symmetry axis and the alignment imposed by the electric field, this behavior cannot be attributed to misalignment of the metallic hemisphere relative to the substrate. This was confirmed by measuring the alignment angle ($\alpha$) of the metallic hemisphere as a function of Θ, both with and without the electric field (see **Methods**). The results, presented in **Supplementary Fig. 4**, show that $\alpha$ remains nearly constant under an electric field, regardless of Θ. Therefore, the observed effect may instead result from small rotations of the JP about the symmetry axis normal to the interface plane. Such rotations could cause variations in propulsion efficiency due to inhomogeneities in the Au coating or local contaminants that interfere at specific angles. To implement closed-loop control while accounting for these effects, the inclination angle Θ was calibrated for each microrobot and frequency regime by manually identifying the near optimal angle prior to a control session. Additionally, because $\vec{B}_{xy}$ is not perfectly perpendicular to the electric field-driven velocity $\vec{V}$, and these variations differ among particles, direct calculation of the required magnetic azimuth angle was impractical. Instead, a proportional (P-) controller approach was used. If the velocity vector $\vec{V}$ deviated from the target direction $\vec{d}$ by an error angle $\alpha_{err}$ exceeding a predetermined threshold (1-4°), the orientation was corrected by rotating $\vec{B}_{xy}$ at an angular velocity Ω proportional to $\alpha_{err}$. The controller was tuned using a single parameter $K_P$, such that $\Omega = K_p * \alpha_{err}$. For a control rate of 20 Hz, a typical value was $K_p$=0.1, corresponding to a rotation rate of 2°/sec per degree of error. This approach produced trajectories with gently curved turns rather than sharp corners, with overshoots that increased with velocity but decreased at higher frame rates. Despite these overshoots, the accuracy achieved was sufficient for the objectives of this study and could be readily improved with a PID controller for finer control. This control strategy is particularly advantageous for selective cargo transport tasks (e.g. synthetic[43] and biological[44] cargo), as the electric propulsion couples with dielectrophoretic (DEP) trapping mechanisms, either positive or negative DEP, depending on the AC frequency.

**Fig. 2C** illustrates an interesting navigation method where magnetic rolling is restricted to the horizontal axis, combined with frequency-controlled orthogonal electric mobility. Unlike the steering approach in **Fig. 2A,B**, this method simplifies motion control by using discrete movements in four directions (left, right, up and down). This approach highlights the alternation between propulsion modes and enables implementation with minimal hardware: a single permanent magnet mounted on a DC motor for rotation around a fixed axis and a function generator with three interchangeable outputs (see example in **Supplementary Fig. 5**). For horizontal motion, the magnetic rotation frequency $\omega$ was fixed at $\omega_{fast}$=5 Hz (300 rpm) in alternating directions, while a high-frequency electric field (5 MHz) suppressed vertical electrical mobility, eliminating the need to toggle the voltage between horizontal and vertical modes. For vertical motion, upward and downward mobility was achieved by selecting the appropriate AC frequency for ICEP and sDEP, with the magnetic field remaining active and rotating at a slower rate of $\omega_{slow}$=0.5 Hz (30 rpm) in the last horizontal direction.

Keeping the magnetic field active is crucial to maintain the vertical alignment of electric propulsion, as random orientation shifts tend to occur without it. Additionally, this approach reliably simulates a setup using a permanent magnet, which cannot be deactivated. The more intuitive approach might be to simply stop the magnetic field rotation and hold it at its settled position. However, this would result in a random magnetic field inclination angle, often hindering electric propulsion due to the alignment sensitivity previously described. Instead, slow rotation ensures efficient propulsion by continuously sampling optimal inclination angles, albeit at the cost of slight diagonal motion instead of a purely vertical trajectory. Furthermore, slow rotation often helps suppress immobilization of the JPs caused by adsorption, electrostatic attraction, or interactions with surface contaminants. Whether ICEP or sDEP is used to achieve upward or downward motion depends on the JP's initial metallic side orientation, which is set at the start of the control signal and may flip during navigation. To address this without requiring continuous orientation tracking, we implemented a dynamic calibration method. By measuring vertical velocity, the controller dynamically switches between ICEP and sDEP if the observed direction of motion opposes the expected one, updating this state in memory for the remainder of the session.

The closed-loop algorithm measures the horizontal ($\Delta x$) and vertical ($\Delta y$) distances to the target and prioritizes motion to reduce the larger error. To prevent diagonal drifts, the algorithm reorients the trajectory using discrete adjustments towards the target line connecting the current and previous waypoint if the distance from that line exceeds a predefined threshold. This control approach results in a characteristic 'staircase' motion, sometimes even along purely horizontal paths due to minor drifts from a straight trajectory. Despite this behavior, the algorithm delivers comparable performance in accurately following complex paths, with the advantage of a highly minimalistic implementation.

Additionally, since the JP maintains an approximately constant orientation, with the metallic side facing up / down (or left / right if horizontal and vertical roles are swapped), it offers the ability to approach samples or cargo from one of four directions. This allows the JP to approach the target with either the apex of one of its hemispheres or its interface region, each offering distinct interaction possibilities.

*Closed-Loop Repeatability vs. Open-Loop Response of Different JPs*

While our closed-loop control system targets a single active particle, other non-controlled particles in the area are also exposed to the same magnetic and electric fields. These particles are passively influenced by the control signals but exhibit diverse responses in velocity, directionality and propulsion efficiency, resulting in unpredictable trajectories. This variability is primarily attributed to the particle's inhomogeneity, including size distribution, fabrication inconsistencies, and the accumulation of debris or microscopic contaminants on individual particles. Additional factors include variations in initial particle orientations and differing local particle-surface interactions, where particles may become temporarily stuck or abruptly change direction after encountering a contaminant.

We leverage this behavior to evaluate the robustness of various closed-loop control strategies compared to an uncontrolled open-loop response within a single experiment. **Fig. 3** demonstrate several examples of such experiments employing different control strategies, in which rectangular or triangular paths were repeated 3-4 times. In every scenario, the controlled active particle consistently maintained its trajectory, passing through similar points in each iteration. In contrast, the uncontrolled particle exhibited trajectories that were shifted, scaled, or rotated.

The observed deviations in each case can be attributed to the specific control strategy and its associated parameters. For example, in hybrid magnetic rolling, small differences in particle size result in varying linear velocities for the same rotation frequency. Additionally, debris or defects on individual particles can alter their response to electric field alignment, resulting in either increased or decreased motion. Similarly, in electric propulsion, variations in initial particle orientations produce rotated trajectories, while differences in optimal inclination angles can result in significantly different velocities. In **Fig. 3B**, the slight rotation of the active particle's trajectory was due to overshoots caused by suboptimal preset parameters, including an insufficient combination of proportional constant ($K_P$=0.09) and frame rate (10 fps) during this session. These results confirm that our closed-loop implementations are adaptable and reliable for programmed navigation tasks, whereas open-loop control sequences often fail to achieve consistent outcomes. Furthermore, the findings underscore key challenges in controlling multiple identical microrobots simultaneously. These include the inevitable cross-responses to shared fields and inconsistencies in the trajectories of neighboring particles, emphasizing that only the active particle's response remains predictable and robust under closed-loop control.

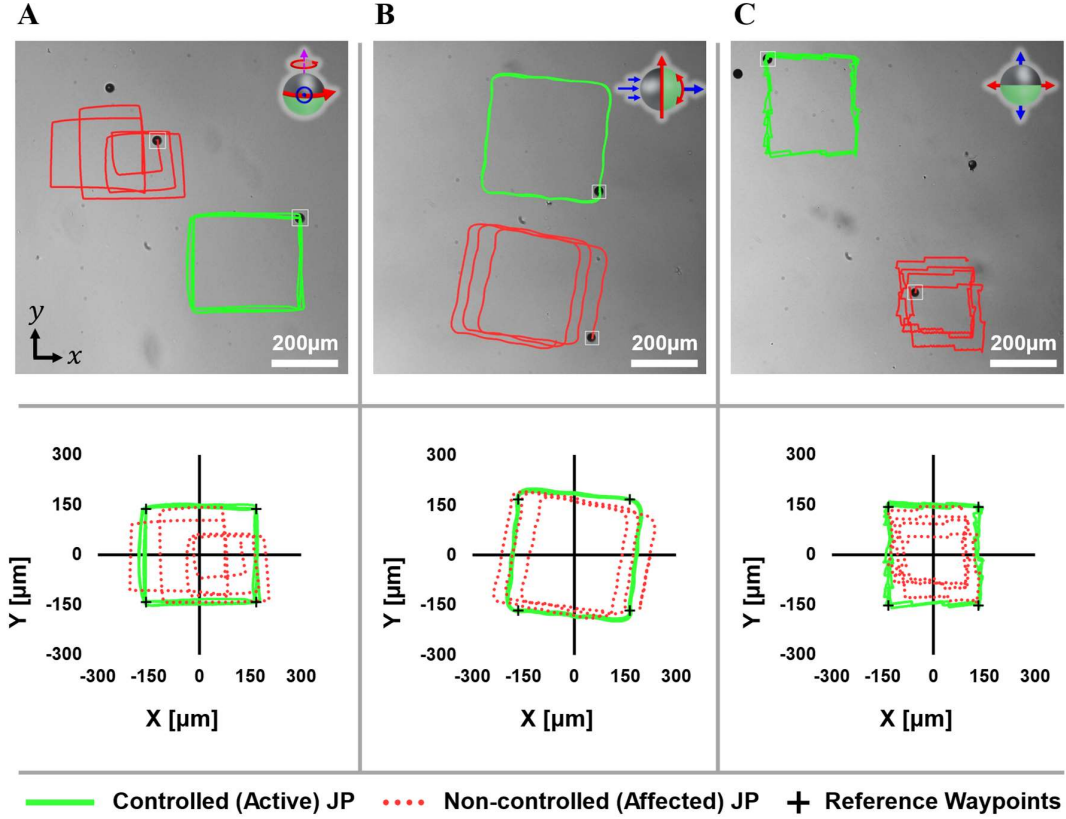

**Fig. 3. Repeated closed-loop patterns – controlled vs. uncontrolled (affected) JPs.** Recorded closed-loop trajectories of 27 μm JPs over 3-4 repetitions of a rectangular reference path defined by four waypoints. Each session involved two tracked particles: the controlled particle (green line) and an uncontrolled particle affected by the same applied fields (red line). The trajectories illustrate the repeatability of the control over a single active particle while highlighting the variability in the response of different particles to the same conditions. The bottom row shows the trajectories in a centered reference system compared with the defined waypoints. Various hybrid control strategies were tested: **A** Magnetic rolling (3 mT, 6-20 Hz) with a high-frequency electric field (5 MHz, 10 $V_{pp}$ - activated during the last two rounds), **B** Electric propulsion (ICEP at 1.5 kHz, 12 $V_{pp}$) combined with magnetic (2 mT) steering, and **C** Orthogonal 4-Directions motion using magnetic rolling (3 mT) at $\omega_{fast}$=5 Hz with an electric field at 5 MHz, 12 $V_{pp}$ during horizontal motion, and $\omega_{slow}$=0.5 Hz with electric propulsion at 1.5 kHz for ICEP and 20 kHz for sDEP for vertical motion, both at 12 $V_{pp}$.

*Magnetic and Electric Field Based 2.5-Dimesnional Manipulation of JPs*

Here, we employ magnetic field gradients to levitate JPs and utilize electrostatic trapping at the chamber ceiling to hold them against gravity, enabling diverse 2.5-dimensional (2.5D) motion. While magnetic field gradients have been employed for 3D manipulation of magnetized objects in viscous liquids[45–47], the relatively weak magnetization of our JPs, induced by a thin nickel (Ni) layer, cannot generate sufficient forces for significant horizontal (XY) translation in addition to vertical (Z) levitation or overcoming surface friction. This limitation is particularly evident when operating with distant magnetic sources over a large workspace, especially with electromagnets, due to the practical limits on magnetic field strength and operational duration. Thus, we use the term '2.5D' to describe movement in 3D space that remains partially constrained to discrete planes or surfaces, yet unlocks new motion capabilities.

**Fig. 4A** depicts the dynamic lifting process through magnetic levitation, ending with attachment to the upper plane. Initially, the microrobot rests in equilibrium on the chamber floor, with the heavier, metal-coated hemisphere facing downward. At t=0, the magnetic field and gradient are activated, both oriented at the +Z direction, similar to the effect of a permanent magnet or a single electromagnet setup operating from above. The metallic side is instantly aligned by the magnetic torque, while the JP begins its upward translation driven by the magnetophoretic forces[48,49]. To generate sufficient force to overcome gravity, relatively high field magnitudes are required, demanding high coil currents.

Typical values in our setup were ~15 mT for the field and ~2000 mT/m for the gradient, when lifting 10 μm JPs. Heavier 27 μm JPs demanded ~17.5 mT for the field and ~2500 mT/m for the gradient, to achieve lifting within reasonable intervals. This is mainly due to the high ratio between the non-magnetic core mass of the particle and the thin Ni layer coating, which is also weakly magnetized. However, greater lifting forces can be generated using powerful NdFeB magnets placed near the sample. During ascent, the microrobot shifts out of the image's focal point, gradually appearing larger and blurrier. Upon reaching the top, as indicated by stabilized focus, the AC field is activated, generating an electrostatic attraction force that holds the JP against its weight once the magnetic field or gradient are deactivated. At this point, the metallic side also aligns due to the electrostatic torque acting on the Au layer.

**Fig. 4B** illustrates the concept of 2.5D trajectories based on this mechanism, using a simple rectangular path. In this example, the microrobot is actuated on the bottom plane along the first two edges, lifted and held at the top plane to traverse the remaining edges, and then descends back to the starting point once the electrostatic force is disengaged. The realization of such a trajectory with manual control is demonstrated in **Fig. 4C-E**, showcasing various modes of actuation: magnetic rolling, assisted by an electric field for better alignment, enhanced friction and ceiling trapping (**Fig. 4C**); ICEP electrokinetic propulsion guided by magnetic steering (**Fig. 4D**); and the same concept with sDEP as the propulsion mechanism (**Fig. 4E**). An interesting effect to account for with magnetic rolling at the top is the inversion of the rotation axis, as the rolling surface is on the opposite side relative to the microrobot. It is also important to note that while the magnetic actuation setup used here technically enables superimposed magnetic field rotation with upward lifting gradients for inverted rolling, such an implementation requires high coil currents for prolonged periods and could be unstable due to the rapid shifting in these high currents, stretching hardware limitations. Our approach, relying on the electrostatic trapping, avoids the need for these strong sustained gradient fields and allows us to use only low-magnitude uniform fields for rotation.

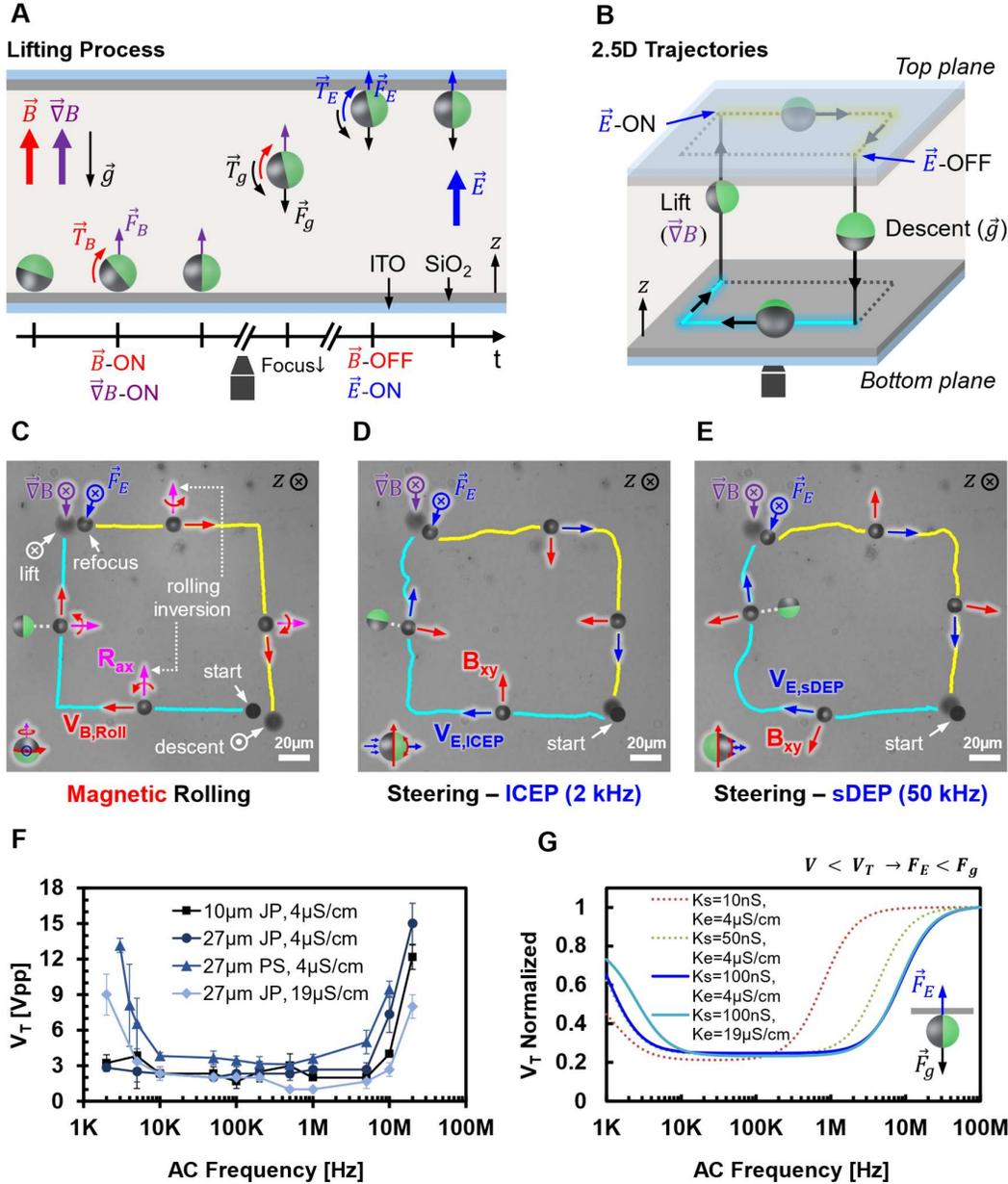

**Fig. 4. 2.5-Dimensional Janus particle manipulation through magnetic lifting and electrostatic trapping**. **A** Sequential process of JP lifting within the ITO chamber using magnetic field gradients, followed by attachment to the chamber ceiling via electrostatic attraction. **B** Illustration of a simple rectangular 2.5D trajectory, with half covered on the bottom plane (cyan line) and the other half on the top plane (yellow line). **C-E** Manually controlled execution of the 2.5D rectangular trajectory using different control methods: magnetic rolling (6 Hz) with a high-frequency electric field (5 MHz, 12 $V_{pp}$) on the top surface, magnetic steering with ICEP (2 kHz, 12 $V_{pp}$), and magnetic steering with sDEP (50 kHz, 12 $V_{pp}$), respectively. All with a 3.5 mT magnetic field. The Janus particle (10 μm in diameter) momentarily leaves the image focus during lifting or descent, requiring adjustments to maintain visibility during the session. **F** Experimental measurements of the threshold voltage ($V_T$) required to hold JPs at the top plane as a function of applied AC frequency. Measurements for bare Polystyrene (PS) particles

are included for comparison. **G** Theoretical model plot of the normalized threshold voltage sweep, based on values of $K_e$ = 4 µS cm$^{-1}$ or 19 µS cm$^{-1}$ for the solution bulk conductivity and $K_s$ = 10 nS, 50 nS, 100 nS for the surface conductance (details in **Supplementary Note 1**).

To explore the limits of the electrostatic trapping mechanism, we thoroughly investigated its capability to hold the particle at the ceiling against gravity. Varying the AC frequencies, we performed a voltage scan, from higher to lower voltage, to determine the threshold *voltage* at which the JPs escape the electrostatic trap and fall down. The particles were not intentionally propelled during this scan, yet responded with different levels of motion depending on the applied frequency. In most cases, near the threshold voltage the propulsion was very weak, with the velocity approaching zero. The results, depicted in **Fig. 4F**, can be directly translated to the imposed uniform electric field magnitude if divided by the chamber height of ~120 µm.

In a low solution conductivity of 4 µS cm$^{-1}$ (mainly used in this study), for JPs with diameters of 10 and 27 µm, the threshold voltage remained at a minimum plateau in most of the examined frequency range beyond ~2 kHz, before rising significantly at high frequencies around 10 MHz, near the upper limit of our measurement range (20 MHz). When comparing to bare polystyrene (PS) particles in the same solution conductivity, we observed a significant rise in the threshold voltage for the PS particles also at low frequencies below 10 kHz. We suspected that this rise resulted from weaker electrostatic attraction forces due to the smaller induced dipole within the PS hemisphere compared to the Au-coated hemisphere, revealing an attenuation of these forces in general at lower frequencies.

The response was qualitatively captured by our theoretical model, plotted in **Fig. 4G**, based on the equivalent dipole of a metallo-dielectric Janus particle[50,51]. This model considers the RC frequency of the induced electric double layer[52], electrode screening[53], and the Maxwell-Wagner high-frequency dispersion[54] (details in **Supplementary Note 1**). We used the surface conductance $K_s$ as a fitting parameter in the model, with the value of 100 nS showing the best correlation to the experimental data. This relatively large value compared to previous experimental estimations (e.g. $K_s$=2.3 nS for 4.8 µm polystyrene beads[55]) may be attributed to additional effects not accounted for in our model, such as the particle-wall interactions. Therefore, $K_s$ is used solely as an effective parameter. Interestingly, at an increased conductivity of 19 µS cm$^{-1}$, the 27 µm JPs exhibited a higher threshold voltage in the lower frequency range, resembling the threshold of bare PS particles – a trend supported by our theoretical model.

Additionally, we measured the effect of magnetic rolling speed on the threshold voltage, expecting a clear trend of upward shift with increased rotation rate, due to the generated hydrodynamic lift force[56–58]. We scanned different magnetic field rotation rates under fixed AC field frequencies of 750 kHz and 5 MHz, which are the edges of a range in which electric propulsion is minimized and the threshold voltage without rolling remained similar. However, compared to the idle state, we found only a small shift of 1-2 Volts at lower rotation rates, which levels off at around 10 Hz (**Supplementary Fig.**

**6**). This shift is relatively minor compared to the observed differences in varying AC frequencies, which could reach up to 12 Volts. These findings suggest that the contribution of the rolling-driven orthogonal force reaches a certain peak, beyond which it is counteracted by the electrostatic force. The initial shift at lower rates may also be associated with increased disturbances to the electrostatic equilibrium, statistically raising the likelihood of detachment.

*Interplanar Navigation and Obstacle Crossing*

Relying on the concept of 2.5D motion via ceiling attachment and combining it with our open-loop and closed-loop control systems, we introduce a variety of new capabilities. These include programmed complex trajectories with multiple transitions between the bottom and the top substrates, facilitated interplanar transitions into elevated surfaces, and obstacle crossing in 3D space. Several such examples are presented in **Fig. 5**. **Fig. 5A** illustrates an interplanar closed-loop navigation session with magnetic rolling, spelling the initials 'TAU', where the letter 'A' segment occurs on the chamber ceiling. Starting on the bottom surface, this trajectory involves a single transition into each plane of motion, with automated application and deactivation of the required magnetic gradient and electric field at each phase. It also inherently includes electrically enhanced rolling motion on the top plane, as it is needed for the electrostatic trapping. **Fig. 5B** presents a similar control session, demonstrating how automated interplanar transitions can be harnessed to produce discrete, separated patterns in one of the planes. This could be practically beneficial in scenarios where the traveling microrobot induces an effect on the substrate (e.g. "writing"[59], localized stimulation[10,60,61]), enabling specific, clearly separated interventions at desired locations, while avoiding unnecessary interactions and preserving other areas from being affected. Furthermore, guiding microrobots to specific positions on the upper substrate and releasing them on-demand mimics the operation of a "pick-and-place" manipulator, with potential applications in cargo transport between compartments[62] and the assembly of microstructures[63,64]. Accomplishing these types of trajectories in a fully closed-loop manner would require knowing the exact moment the microrobot reaches the top surface when being lifted, and the moment it reaches the floor after descending. This could likely be achieved with a sequence of calibrations and image processing methods. However, due to the complexity of this task outweighing the benefits for our purposes, it was avoided in this study. Instead, we employed a more convenient approach, adding one open-loop component to our control algorithm while keeping it fully automated. By measuring the approximate time it takes for a JP to be lifted and the time it takes for it to descend, we simply added an adjustable standby phase with a few extra seconds during each of these events, after which control was resumed, including automated application of the electric field required for attachment. For example, 20-25 seconds for lifting and 15-20 seconds for descending were typical values in our setup for 10 μm particles, while 27 μm particles generally required shorter intervals of 10-20 seconds to lift and 3-4 seconds to descend.

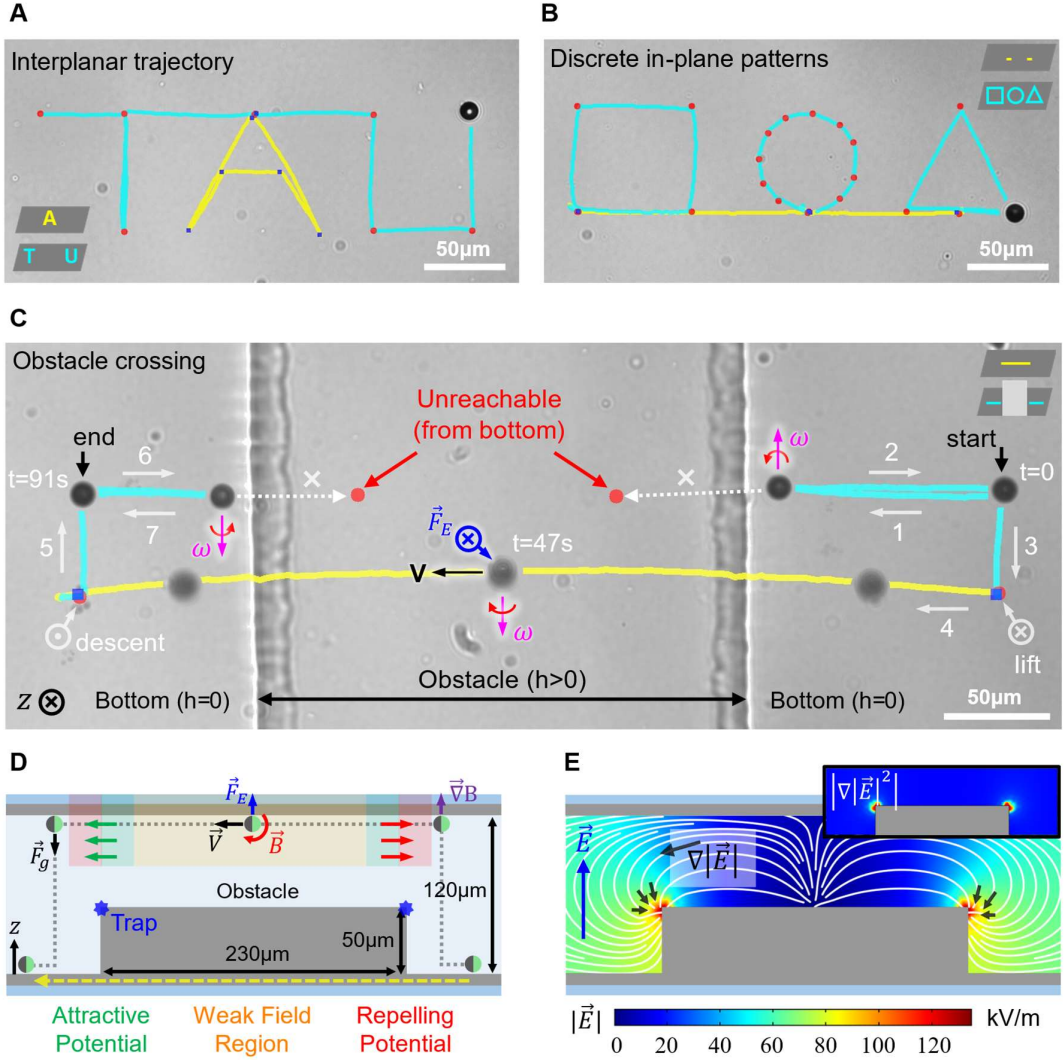

**Fig. 5. 2.5-Dimensional closed-loop control and obstacle crossing in 3D space**. **A** Interplanar trajectory demonstrating the initials "TAU", where the letter 'A' (yellow) is formed on the chamber ceiling. Bottom-plane waypoints are indicated by red circles, and top-plane waypoints are indicated by blue squares. A 10 μm in diameter JP was used in these experiments. **B** Discrete in-plane patterns of three shapes fully separated on the bottom surface, avoiding continuous intermediate motion. Both (A) and (B) are achieved using magnetic rolling (3.5 mT, 6-12 Hz) and electrostatic trapping on the ceiling (12 $V_{pp}$ in (A) and 10 $V_{pp}$ in (B), both under 5 MHz). **C** Fully automated obstacle-crossing session with magnetic rolling (3.5 mT, 18 Hz) on the ceiling above an SU-8 barrier. The applied AC field was 15 $V_{pp}$, 5 MHz. **D** Illustration of the crossing process, including the obstacle's dimensions. The sharp corners of the obstacle shape the electric field, introducing dielectrophoretic (DEP) forces. These induce a repelling potential when approaching the obstacle, an attractive potential when departing it, and a weaker field in the middle region that may lead to microrobot detachment. Additionally, strong field gradients at the corners can trap the crossing particle. **E** Numerical simulation showing the electric field magnitude ($|\vec{E}|$) and the streamlines of its gradient ($\nabla|\vec{E}|$), along with the magnitude of the vector field $\nabla|\vec{E}|^2$ around the obstacle (inset). The simulated voltage between the top and bottom boundaries was set to 7.5 V DC, approximating the AC

voltage of 15 Vpp. This approximation to DC is valid at high frequencies, beyond the RC time of induced electric double-layer screening.

Theoretical calculations estimated settling times of 6.93 seconds for 10 μm JPs and 1.68 seconds for 27 μm JPs when dropped from the ceiling of the microchamber, accounting for gravity, buoyancy and Stokes' drag (details in **Supplementary Note 2** and **Supplementary Table 1**). However, these calculations assume an idealized model where each metal layer deposited on the particle (Cr-Ni-Au) forms a hemispherical cap with a uniform known thickness. Previous studies have shown that metal deposition on spherical particles is non-uniform, with greater thickness near the particle's apex and tapering towards the edges due to the surface curvature[65]. This non-uniformity leads to a reduced metallic mass compared to the idealized model. Moreover, heterogenous particle sizes and material loss during fabrication, as well as interactions with the surrounding solution and substrate, contribute to discrepancies between theoretical predictions and experimental measurements. Notably, applying the same theoretical analysis to bare polystyrene particles predicts settling times of 40.36 seconds and 4.68 seconds for 10 μm and 27 μm particles, respectively, highlighting that the observed settling times for JPs are within the expected intervals, and are consistent with the added mass of the deposited metals.

**Fig. 5C** displays a closed-loop obstacle crossing session, utilizing magnetic rolling as well, conducted in a specialized chamber containing several U-shaped SU-8 structures (see **Supplementary Fig. 7**). These structures, reaching a uniform height of ~50 μm within the available 120 μm gap, were substantial enough so that a single wall could effectively block direct or lateral passage on the bottom, while permitting crossing from above. Therefore, they were regarded as 3D obstacles, as demonstrated here, or as elevated surfaces for descending upon and traveling along the structure itself. Additionally, their transparency enabled continuous visibility of the microrobot when positioned above.

The crossing trajectory began on the bottom plane, with the first waypoint placed within the obstacle's borders, emphasizing that the microrobot could not reach it from below due to collisions with the wall. We then manually switched to the next waypoint to continue with the programmed path, lifting the microrobot to the ceiling and approaching the obstacle again, but this time successfully crossing it from above. Upon reaching the opposite side, the microrobot was automatically lowered, with the electric field deactivated. Before finally settling at the target destination, another unreachable waypoint was tested, reaffirming that crossing from below was impossible. During this session, the image focus was set at an intermediate level between the bottom and the top surfaces, minimizing visual distortions caused by the obstacle's edges and ensuring continuous, reliable tracking.

Throughout our attempts to achieve successful crossing, several challenges and effects were observed, as shown in **Fig. 5D**. When approaching the obstacle from the top substrate, the microrobot encountered resistance from a force that slowed its motion and pushed it back when propulsion was halted. Upon entering the middle region directly above the obstacle, the microrobot could lose attachment to the ceiling, causing it to fall onto the surface. After overcoming these areas and successfully crossing, the microrobot accelerated in velocity as it moved away from the obstacle region, mirroring the resistive force encountered during the initial approach. These phenomena arise from electrokinetic effects caused by the presence of the obstacle structure under the applied electric field, as revealed by the numerical simulation shown in **Fig. 5E**.

The obstacle, made of a dielectric material with two sharp corners at its edges, locally deforms the electric field induced by the voltage difference between the parallel ITO slides. This deformation causes the field to bypass the sides of the obstacle, reducing its strength in the middle region while amplifying it near the edges, resulting in significant electric field gradients. Given the low conductivity solution and high AC frequencies used (500 kHz - 5 MHz), this effect generates positive dielectrophoretic (pDEP) forces on the JP[66], oriented toward the local maxima of the electric field gradient[67,68]. The DEP forces induce motion along the gradient streamlines near the top, creating the observed repulsion and attraction effects. In some cases, the forces are strong enough to overcome the electrostatic attraction at the top and trap the JP at one of the corners. However, this was avoided with trial and error by applying higher voltages and moving the JP more quickly past the edges.

*3D Cargo Transport and Operation at Higher Solution Conductivities*

Here we leverage interplanar transitions to demonstrate a potential application of the hybrid JP microrobot as a 3D "pick-and-place" manipulator, capable of carrying cargo between distinct regions with intermediate motion on the top surface. Specifically, we demonstrate cargo transport between two compartments separated by a virtual wall (see green line in **Fig. 6A**) which also simulates a physical boundary (e.g. a wall) that cannot be crossed from the bottom surface. Both synthetic (1.1 μm diameter polystyrene microspheres) and biological (living *E. Coli* bacteria) cargo was used in two separate sessions of the task, presented in **Fig. 6**. In both cases, the JP was actuated via magnetic rolling, while the trapping and release of cargo was controlled by generation and deactivation of dielectrophoretic (DEP) forces, resulting from the local electric field gradients induced by the JP in response to the applied external AC field[8,43].

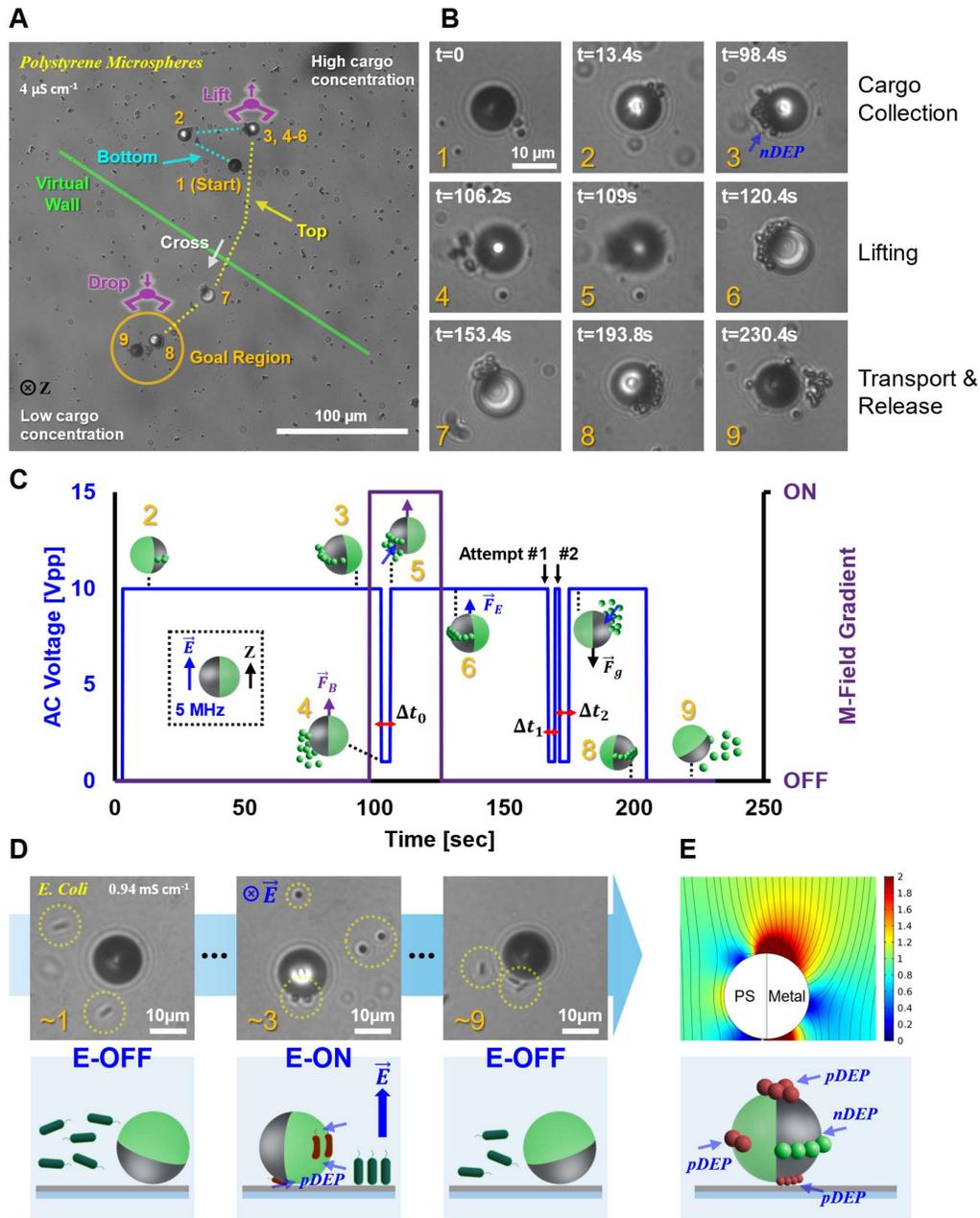

**Fig. 6. Interplanar transport of synthetic and biological cargo with DEP trapping and magnetic rolling**. **A** Timelapse image of a 10 μm diameter JP carrying 1.1 μm diameter PS microspheres from a high- to low PS concentration region above a virtual wall marker, which simulates a physical barrier on the bottom. **B** Zoom-in to key events corresponding to the position numbers noted in (**A**), including cargo collection, lifting of the microrobot with the cargo, and transport on the ceiling to the goal region where the cargo is released. The high-frequency (5 MHz) electric field induces a negative dielectrophoretic (nDEP) force on the PS microspheres[43], attracting them to the metallic hemisphere of the JP. The cargo remains trapped during magnetic rolling (3.5 mT, 1.5 Hz) and is only released when the electric field is deactivated. **C** Plot of the applied AC (5 MHz) voltage and magnetic field gradient (2500 mT/m at 17.5 mT field magnitude) vs. time, with side-view illustrations of the JP-cargo interactions.. Lowering and restoration of the AC voltage (from 10 $V_{pp}$ to 1 $V_{pp}$ and back) to release the JP

from the surface upon lift or descent was performed at intervals of $\Delta t_0 \approx 3.8$ s, $\Delta t_1 \approx 2.8$ s, and $\Delta t_2 \approx 3.8$ s. Since $\Delta t_1$ was too short to disengage from the ceiling, the JP was re-attracted to the top with the cargo, before successful detachment with $\Delta t_2$.. **D** A similar task was performed with live *E. Coli* bacteria as biological cargo, with their viability retained throughout the process, as evidenced by their motility upon release. Selected images corresponding to similar events in (**B**) are shown along with illustrations of the instantaneous electric-field-based JP – bacteria interactions[10].**E** Top - numerically computed electric field magnitude (normalized by the applied uniform electric field) and streamlines around a 10 μm JP positioned 300 nm above the surface between two parallel electrodes distanced 120 μm from each other[18]. Bottom – schematic of distinct cargo trapping positions on the JP via pDEP and nDEP, corresponding to the local maximum and minimum electric field intensity, respectively[43].

The coupling of the cargo trapping onto the JP with the electric field requires accounting for the electrostatic force that attracts the JP to either the bottom or top surfaces during the initiation of lift or descent sequences when transitioning between planes. To enable these transitions without losing the trapped cargo, we significantly lowered the AC voltage – just enough to release the JP from the surface while still maintaining cargo trapping. Although some cargo temporarily detached, it was quickly recovered by restoring the voltage. Hence, the timing of these voltage shifts was crucial – it had to be long enough for complete disengagement of the JP from the surface, yet short enough to retain the cargo. **Figures 6B-C** illustrate this process, showing the separation and reattachment of polystyrene (PS) microspheres to the carrier JP during the lifting sequence, in correlation with the applied magnetic and electric fields.

Depending on the solution conductivity and electric field frequency, different types of cargo may exhibit positive DEP (pDEP) or negative DEP (nDEP) response, leading to accumulation in distinct regions of electric field maximum or minimum on the JP surface, respectively[43,69]. In previous studies[10,43], we have shown that pDEP-responsive cargo is predominantly trapped along the metallo-dielectric interface of the JP or at the equator of its PS hemisphere, while nDEP-responsive cargo is mostly trapped at the equator of the metal-coated hemisphere (see **Fig. 6E**). Additionally, we have previously[18] shown that magnetic rotation induces very strong hydrodynamic shearing forces at the poles of the JP. Therefore, to minimize shear-induced detachment, we preferred using nDEP for efficient trapping at the metal hemisphere's equator, where such forces are reduced. Our synthetic cargo particles exhibited nDEP response in a solution conductivity of 4 μS cm$^{-1}$ at frequencies $\geq$ 400 kHz (see **Supplementary Fig. 8**), allowing us to perform the task in an environment consistent with most other experiments in this study. However, live *E. Coli* bacteria were previously found to exhibit only pDEP response in the frequency range of $10^3$ - $10^7$ Hz under such low-conductivity solutions, with the transition from pDEP to nDEP response occurring between 0.1 mS cm$^{-1}$ and 10 mS cm$^{-1}$ conductivity[70]. We therefore first examined several relatively high-conductivity solutions of diluted PBS to facilitate bacteria transport[66]. At $\approx$1.83 mS cm$^{-1}$ (PBS diluted x10), the highest available hardware (Agilent 33250A) output without amplifier (20 $V_{pp}$) was insufficient to stably hold the 10 μm JP at the top surface when examined at 1 MHz and 5 MHz frequencies. This may result from increased induced-charge electric double layer (EDL) screening of the metallic hemisphere in higher conductivities[18], and might as well be associated with the additional contribution from electrothermal convection, which scales with solution conductivity[71]. However, at ≈0.94 mS cm$^{-1}$ (PBS diluted x20), stable electrostatic trapping was achieved at ~7 $V_{pp}$ or above at 5 MHz, making this condition suitable for interplanar *E. Coli* transport. Notably, the solution did not include Tween-20 as a surfactant to prevent damaging the bacteria, hence the JP adhered to the surface more frequently during navigation, resulting in longer time for completion.

**Fig. 6D** illustrates part of the *E. Coli* transport process in the 0.94 mS cm$^{-1}$ PBS solution, showing the vertical alignment of the bacteria due to electro-orientation when the electric field was activated, and their observed trapping locations on the JP. The occurrence of bacteria trapping on the PS hemisphere and in the gap between the metal hemisphere and the substrate indicates an exclusive pDEP response, suggesting that the solution conductivity was still insufficiently large to induce nDEP trapping of *E. Coli*. Nevertheless, several bacteria were successfully transported and released, as pDEP trapping at the PS hemisphere minimized hydrodynamic shearing similar to nDEP trapping of the PS particles at the metallic hemisphere[18]. Notably, the bacteria remained viable, as evidenced by their motility after release.

**Discussion**

We have methodically integrated versatile hybrid magnetic and electric actuation mechanisms for metallo/magneto-dielectric Janus particles (JPs), unlocking capabilities that surpass the limitations of traditional surface-bound motion. By designing and characterizing a set of motion control strategies within a single plane, drawing on both established and novel approaches, we have laid a robust foundation for dynamic and precise particle manipulation. Building upon this groundwork, we introduced a higher level of motion, featuring interplanar transitions and 2.5-dimensional (2.5D) trajectories. This advancement, enabled by magnetic levitation and electrostatic trapping, extends the mobility of JPs into 3D space beyond planar constraints. Furthermore, the incorporation of the automated navigation system developed in this work exemplifies the transformation of JPs from simple micromotors to multifunctional microrobots.

The presented capabilities open new pathways for practical applications of our JP microrobots and other potential designs of hybrid magnetic and electric microrobots in general. Interplanar traversal enables microrobots to overcome obstacles and transition between separate compartments, paving the way for tasks such as "pick-and-place" operations which were explored here for both synthetic and biological cargo. Discrete surface patterning with precise closed-loop control can allow microrobots to make marks or deliver stimuli exclusively at designated locations, ensuring that unaffected areas remain untouched – a functionality akin to "printing". Additionally, utilizing the ceiling as a transitional plane facilitates unique trajectories, enabling microrobots to navigate 3D environments by descending to lower levels as needed. This capability also supports targeted deposition of particles above selected areas, potentially aiding in the assembly of 3D structures. These features hold significant promise for biomedical research and microfluidic applications, where selective and precise interventions are paramount.

While our hybrid control system demonstrated robustness and repeatability, certain operational considerations arise with electric actuation due to electrokinetic phenomena. For instance, the requirement for low-conductivity solutions to support electric propulsion limits the environments where full utilization of the hybrid microrobot capabilities is possible. However, transitioning to magnetic rolling can bridge this gap in setups with varying media, where some regions meet the necessary conductivity conditions[12]. Furthermore, we have shown here and in our previous studies that electric fields can still cooperate with magnetic rolling for actuation, alignment and functionality in high-conductivity solutions[18]. Another important consideration is the impact of dielectrophoretic (DEP) forces resulting from local field distortions near objects, particularly with sharp edges that intensify DEP forces, which can influence microrobot behaviour and complicate control. We addressed this challenge during the obstacle-crossing experiments by modelling the environment, characterizing the effects and the limitations, and tweaking control parameters to successfully achieve the goal of crossing. Future work could focus on further optimizing actuation strategies to improve adaptability to spatially and temporally varying conditions (e.g. using PID, model-predictive, and more advanced controllers), ensuring reliable performance across diverse scenarios. Additionally, operational efficiency could be improved by incorporating stronger electromagnets positioned exclusively at the top and bottom of the chamber for selective activation during lifting and descent phases, significantly reducing the standby time required for these transitions. Our work highlights the transformative potential of hybrid control systems in overcoming motion control limitations for microrobots, enhancing their mobility and range of tasks. The introduction of 2.5D manipulation lays the foundation for innovative microrobotic applications beyond a single surface.

**Methods**

**Preparation of Janus particles and Microchamber Systems.** Metallo/magneto-dielectric JPs with diameters of 10 and 27 μm were fabricated by coating fluorescent polystyrene particles (Sigma Aldrich, St. Louis, MO, USA) with 15nm of chromium (Cr), followed by the deposition of 55 nm nickel (Ni) and 15 nm of gold (Au) layers using an electron-beam evaporator (PVD20-EB, Vinch technologies), as described previously[7]. The coated JPs were released via sonication in deionized (DI) water and rinsed three times with DI water before being suspended in a low-conductivity potassium chloride (KCl) solution (~4 μS cm$^{-1}$) with 0.05% Tween-20 (Sigma Aldrich) as a non-ionic surfactant to minimize adhesion to the ITO substrate. The microchamber system consisted of a circular microfluidic chamber formed by a spacer positioned between two indium tin oxide (ITO)-coated glass slides (Delta Technologies), as described in earlier work[7]. To reduce particle adhesion to the substrates, both ITO-coated glass slides were further coated with a 20 nm-thick silicon dioxide ($SiO_2$) layer using a sputter coater (PVD20-S, Vinch technologies). The thin spacer, which defined a 10 mm-diameter microchamber, was constructed from a 120 μm-thick double-sided tape (3M) which was cut using an electronic cutting machine (Silhouette Cameo 3, Silhouette America Inc.). Obstacles within the microchamber were created on the bottom ITO-coated glass slide by fabricating semi-enclosed U-shaped structures with a thickness of ~50 μm. These structures were made using standard photolithography techniques with SU-8 negative photoresist (SU8-2025, MicroChem), as detailed in a previous study[72].

**Experimental Setup**. The microfluidic chip with microchambers was placed on a Nikon TI inverted epi-fluorescent microscope (Nikon, Japan) equipped with an Andor Neo camera (Andor, Belfast, UK). Electric signals were generated using a single-channel function generator (Agilent 33250A, Keysight Technologies, Inc., Santa Rosa, California, U.S.), controlled via a GPIB-USB interface. Sine wave signals with frequencies ranging from 1kHz to 20MHz and peak-to-peak voltages up to 20V were applied. Magnetic actuation was carried out using the MagnebotiX MFG-100-i field generator (MagnebotiX AG, Zurich, Switzerland)[73]. This system consists of an array of eight electromagnets positioned above the sample workspace, capable of producing both magnetic field gradients and gradient-free uniform fields. Gradient fields (15-17.5 mT, 2000-2500 mT/m) were used exclusively for JP levitation, while uniform idle fields were applied for steering, and uniform rotating fields were utilized for magnetic rolling, with a typical magnitude of 3.5 mT and rotation frequencies of up to 50 Hz. Additional details are provided in **Supplementary Fig. 1**.

**Realization of Open- and Closed-loop Control.** The control system was developed using a custom Python program that managed live image capture from the camera and communicated with the hardware via digital output commands. The AC function generator was operated by setting the voltage and frequency parameters and toggling its output state. Similarly, the MagnebotiX field generator was controlled by specifying magnetic field and gradient components, which the system translated into corresponding coil currents. Additionally, acquisition and logging modules recorded experimental data, including the applied magnetic and electric outputs at each time step, generating a comprehensive dataset for post-analysis. For open-loop control, keyboard inputs were used to adjust the magnetic field for steering, rolling and lifting and to toggle the electric field for propulsion through interchangeable control schemes. For closed-loop control, the system tracked the position and velocity of the microrobots using a pipeline that integrated the CSRT tracker from the OpenCV library with additional image processing for precise center-point extraction. A dynamically updated region of interest (ROI) was defined around each tracked microrobot, enabling efficient tracking of multiple particles while actively controlling only the selected one. Navigation was guided by a sequence of interconnected waypoints forming a defined path, either manually marked on the live image feed or sampled from predefined geometric shapes (e.g., rectangles, triangles, lemniscates). Each waypoint included a marker indicating whether it was on the bottom or top substrate, allowing the controller to apply lifting and descent sequences when necessary, and to use an inverse rotation axis when rolling on the top substrate. Each hybrid control strategy was implemented through tailored algorithms applied at every frame update, ensuring synchronization with the camera frame rate (typically around 20 Hz). The control algorithms included tunable parameters dynamically loaded into the system, such as the $K_P$ constant for the steering algorithm and the rotation frequency range for the magnetic rolling algorithm. Given the low Reynolds number environment, where particle inertia is negligible, the system's response to control signals was effectively instantaneous.

**Measurement of Alignment Angles.** To measure alignment angles, 27 μm Janus particles with a fluorescent polystyrene hemisphere were introduced into the microchamber and visualized using an epi-fluorescent microscope with a 20x objective lens. Images were captured with an exposure time of 50 ms. The orientation angle of the JPs' dielectric-metal interface relative to the XY-plane ($\alpha$) was determined by analyzing fluorescence intensity using the following equation[35,74]: $\alpha = 2\sin^{-1}(\sqrt{I^*})$. Here, $I^*$ is the non-dimensional areal integrated intensity around the JP, calculated by normalizing the measured average pixel intensity $\bar{I}$ with the extreme cases of the particle's metallic ($I_{min}$) and dielectric ($I_{max}$) hemispheres facing downward: $I^* = \frac{\bar{I}-I_{min}}{I_{max}-I_{min}}$. Reference measurements of $I_{min}$ and $I_{max}$ were obtained by applying specific magnetic and electric fields to stabilize the particles at known angles in equilibrium. Due to inconsistent intensity measurements in high angles approaching 180°, $I_{max}$ was estimated based on the intensity measured at 90°, using the approximation $I_{max} \cong 2\bar{I}_{90°} - I_{min}$. This method provided reliable angle estimations up to 90°, which was the relevant range for this study. The orientation angle of the JP's interface relative to the x-axis ($\beta$) was measured by applying a binary threshold to isolate the bright hemisphere. The bright contour was then detected, and its primary axis direction was extracted using image moments with the OpenCV library.

**Threshold Voltage Scan.** Janus particles (10 μm and 27 μm) were levitated to the chamber ceiling using magnetic field gradients (2000-2500 mT/m) generated by the MagnebotiX system under field magnitudes of 15-17.5 mT. Once at the top, particles were held against gravity via electrostatic trapping, applying an initial voltage of approximately 12 Vpp at 5 MHz to ensure firm attachment while preventing lateral motion. To determine the threshold voltage, the AC frequency was set to various predetermined values, and the voltage was decreased incrementally by 1 V until particle detachment occurred, indicating that the threshold was one step above the detachment point. For bare polystyrene particles, lifting was achieved by manually flipping the chamber, allowing gravity to bring the particles to the ceiling. The voltage was applied to secure attachment before the chamber was flipped back. In both cases, equilibrium stability was tested by lightly tapping the table, reducing the likelihood of particles adhering to the surface due to local interactions irrespective of the applied voltage. A similar protocol was used to assess the impact of lateral magnetic rolling on threshold voltage, with the magnetic rotation rate fixed each time while maintaining rolling motion throughout the voltage-lowering process. The AC frequency was set to either 750 kHz or 5 MHz in this case.

**Numerical Simulations.** The numerical simulations, performed in COMSOL™ 5.3 to qualitatively analyze the deformation of the electric field between the parallel ITO electrodes caused by the obstacle structure or a JP, used a 2D model. The geometries comprised a rectangular domain representing the space between the electrodes (120 μm height and 1200 μm (Fig. 5E) or 300 μm (Fig. 6E) width), with a rectangle (50 μm height and 230 μm width) subtracted at the lower edge to represent the obstacle (Fig. 5E), or a circle of 10 μm diameter subtracted at 300 nm above the substrate to represent the JP (Fig. 6E). The problem was analyzed under quasi-steady DC conditions, focusing solely on the qualitative shape of the electric field magnitude, as well as its streamlines (Fig. 6E) or its gradient streamlines (Fig. 5E) under high-frequency AC voltage. This approximation is valid at high frequencies, beyond the RC time of the induced electric double layers. Accordingly, a stationary electrostatic model (electric currents) was used, solving the Laplace equation for the electrostatic potential. The boundary conditions included external electric potentials of 0 V (ground) at the lower electrode and 7.5 V (Fig. 5E) or 5 V (Fig. 6E) at the upper electrode. For the dielectric obstacle, electrically insulating boundary conditions were applied at its edges. For the JP, this condition was applied at the dielectric PS hemisphere, while a floating potential condition was applied to the conductive metal hemisphere to ensure zero total charge.

**Preparation of Synthetic and Biological Cargo.** Synthetic 1.1 μm cargo particles (Polystyrene particles, Fluoro-max) were suspended in a ~4μS cm$^{-1}$ KCl solution with 0.05% Tween-20 (Sigma-Aldrich) and introduced into the microfluidic chamber with high concentration near the injection point, creating distinct regions of high and low concentrations within the chamber. *E. Coli* bacteria (MP 289, obtained from Prof. Yechezkel Kashi, Technion) were cultured at 37 °C in LB agar plates and grown as single colonies to serve as a biological cargo. Bacteria were picked up from a colony for experiments using a pipette tip and washed three times with a relatively high-conductivity (≈0.94 mS cm$^{-1}$) x20 diluted PBS (Sigma-Aldrich) solution for the experiment, before being introduced into the microchamber in a similar manner.

**Acknowledgements**

We acknowledge support from the Israel Science Foundation (ISF) (1429/24). We thank Tel-Aviv University nano-center for assisting with the fabrication of the Janus particles. We thank Prof. Yechezkel Kashi from the Faculty of Biotechnology and Food Engineering at the Technion for providing us with the *E-Coli* bacteria.

# Supplementary Materials

# Hybrid Magnetically and Electrically Powered Metallo-Dielectric Janus Microrobots: Enhanced Motion Control and Operation Beyond Planar Limits

Ido Rachbuch[1], Sinwook Park[1], Yuval Katz[1], Touvia Miloh[1], Gilad Yossifon[1,2]

[1]School of Mechanical Engineering, Tel-Aviv University, Israel
[2]Department of Biomedical Engineering, Tel-Aviv University, Israel

**Figures:**

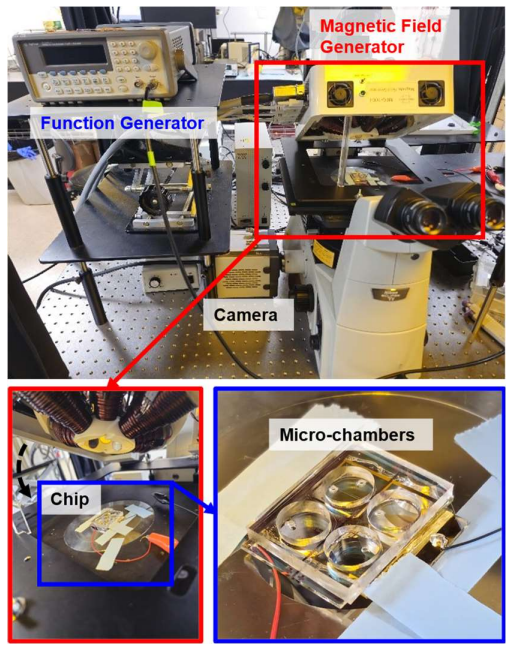

**Supplementary Fig. 1. Images of the experimental setup.** The chip containing micro-chambers is positoned on an inverted microscope for imaging, with a magnetic field generator mounted above it. A function generator is connected to the parallel ITO-coated slides within the chip to apply electric fields. All devices, including the camera, are controlled by a single computer program.

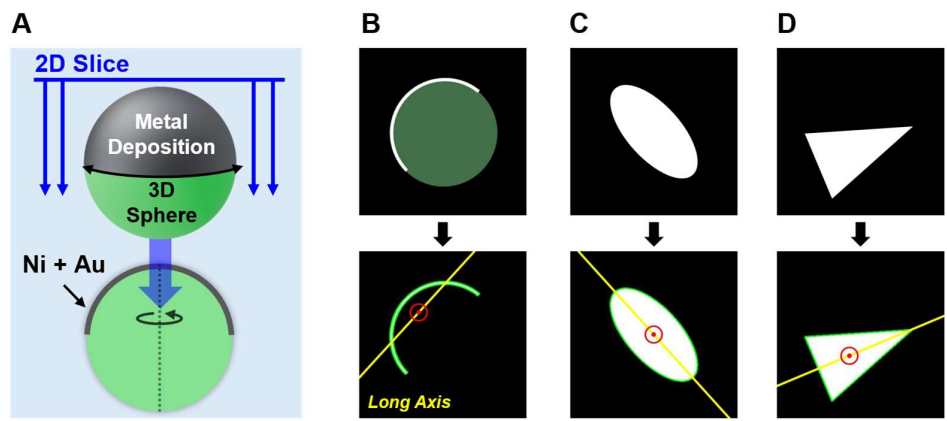

**Supplementary Fig. 2. Long axis of the metallic layers in metallo/magneto-dielectric Janus particles: A heuristic 2D approximation.** **(A)** Diagram illustrating the slicing of a 3D spherical, ideally axisymmetric Janus particle along its middle plane, producing a represnetative rotation-invariant 2D geometry. **(B)** Schematic image of the metallic layer's 2D geometry overlaid with the inner sphere for visualization. Below, the same image is analyzed using contour extraction (green outline) and image moments to determine the centroid (red circular marker) and orientation line (yellow) of the isolated shape, representing the direction of the long axis. **(C)-(D)** Reference analyses of elongated shapes (ellipse and triangle) with clear long axes for comparison. Image processing was performed using the OpenCV library.

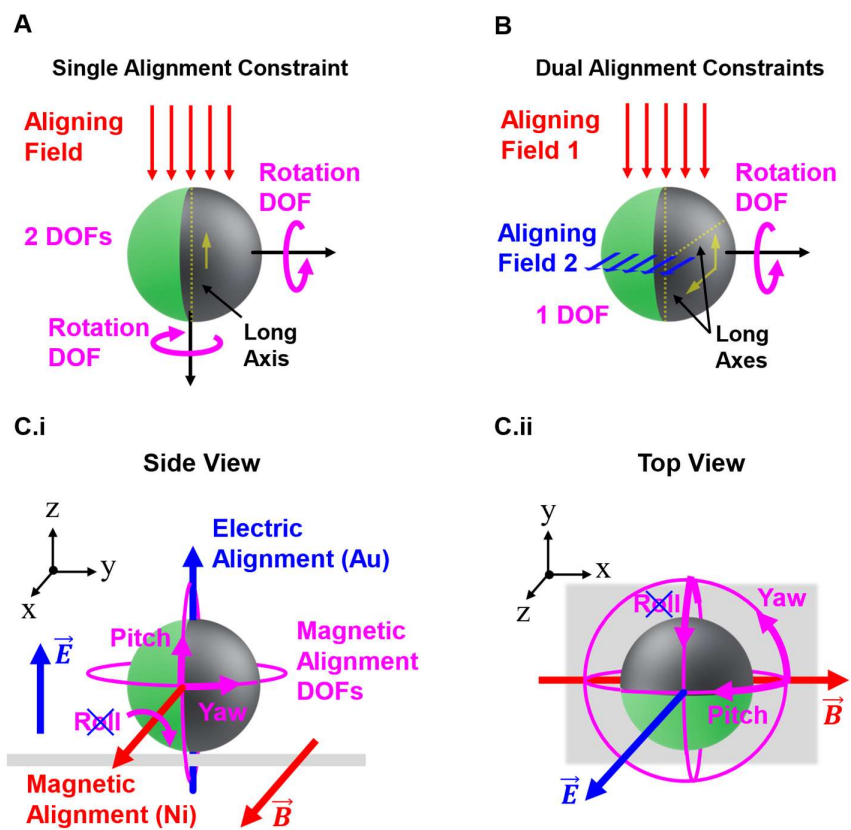

**Supplementary Fig. 3. Rotational alignment constraints of non-magnetized metallo/magneto-dielectric Janus particles (JPs) under external electric and magnetic fields. (A)** Schematic showing the rotational constraint imposed by a single aligning field (magnetic or electric) acting on the metallic hemisphere of the JP. The long axis of the metallic hemisphere aligns with the field to acheive an energetically favorable state of the dipole indcued by the field. This constraint leaves two rotational degrees of freedom (DOFs) around two axes: one parallel to the field direction and the other normal to the JP's interface plane, owing to the gemoetry's rotational symmetry about this axis. **(B)** When two non-parallel and independent fields (electric and magnetic) are applied simultaneously, dual alignemnt constratins are introduced. Consequently, a single rotational DOF remains around the axis orthogonal to both field vectorrs. **(C)** Side view **(i)** and top view **(ii)** projections of a JP positioned on a flat surface, depicting the available DOFs for magnetic alignment under a superimposed electric field normal to the surface. The electric alignment restricts JP rotation to the instantenous z-axis (yaw) and y-axis (pitch), while rotation around the x-axis (roll) is prohibited, leaving two accesible rotational DOFs..

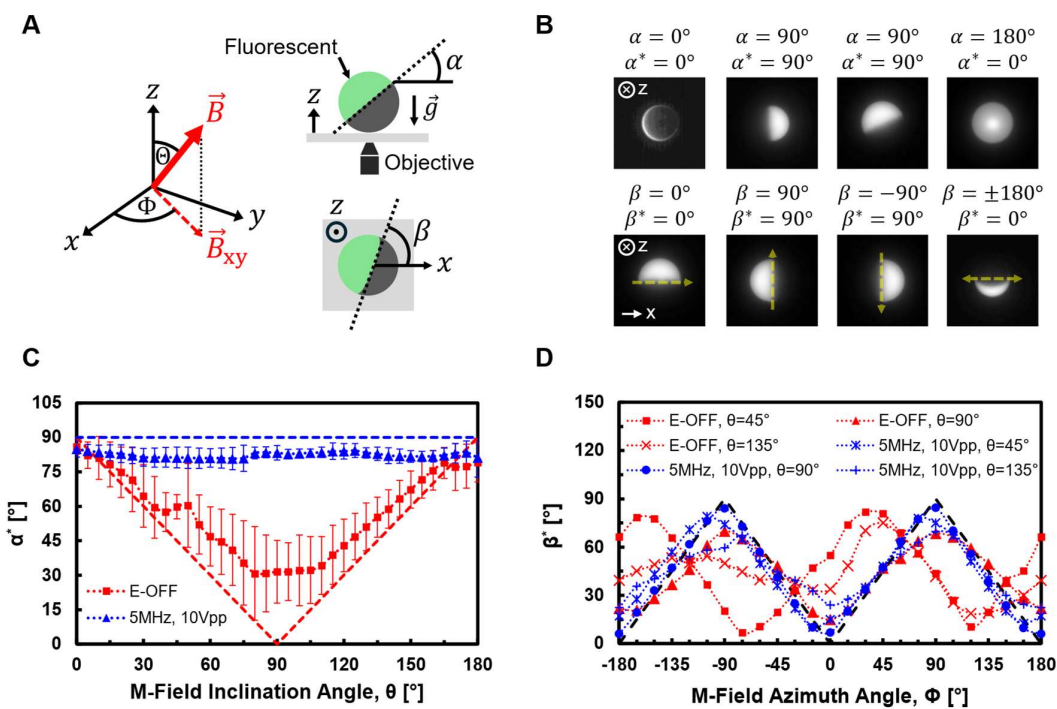

**Supplementary Fig. 4. Alignment angles of non-magnetized metallo/magneto-dielectric Janus particles (JPs) under external magnetic and electric fields.** (A) Schemataic illustration defining the magnetic field direction angles: Θ (inclination) and Φ (azimuth), and the JP orientation angles: $\alpha$ (relative to the XY-plane) and $\beta$ (around the z-axis, relative to the x-axis). (B) Controlled JP orientations acheived by applying specific magnetic and electric fields to stabilize the particles in equilibrium. The bright side corresponds to the fluorescent polystyrene hemisphere, while the dark hemisphere is Ni + Au coated. The effective angles $\alpha^*$ and $\beta^*$ are represented within the range [0, 90]°, corresponding to completely horizontal and completely vertical alignments, accounting for symmetry. Images are normalized to the full grayscale range. (C) Measurements of $\alpha^*$ as a funtion of the magnetic field inclination angle Θ, under a uniform magnetic field (3.5 mT) w/ and w/o an additional electric field (5 MHz, 10 Vpp). Dashed lines represent the ideal theoretical response, neglecting effects such as gravitational torque. (D) Measurements of $\beta^*$ as a function of the magnetic field (3.5 mT) azimuth angle Φ, for inclination angles of 45°, 90° and 135°. The ideal theoretical response is shown as a black dashed line. Standard deviations range from 2° to 46°, with an average of 16°. All measurements were performed with 27 μm JPs under 20x magnification.

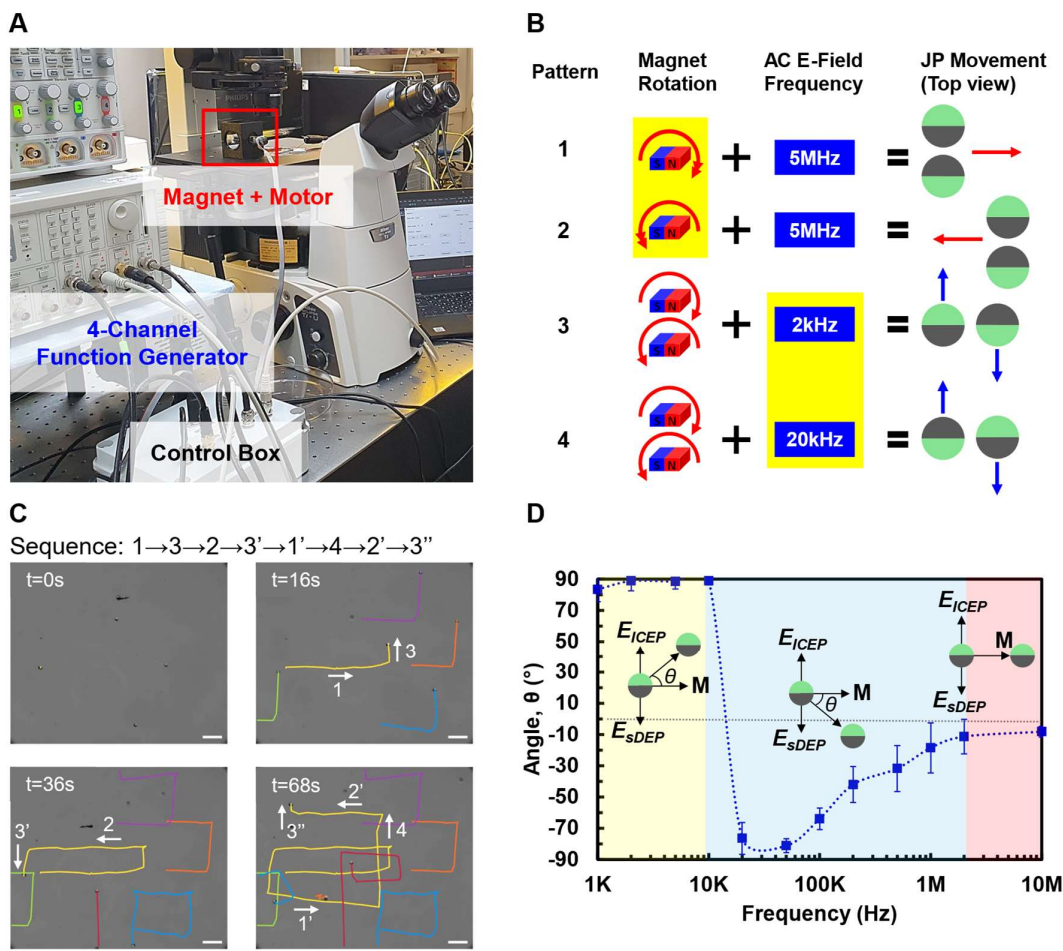

**Supplementary Fig. 5. Orthogonal magnetic rolling and electric propulsion using a rotating permanent magnet.** **(A)** Photograph of the experimental setup, featuring a permanent NdFeB magnet mounted on a DC motor, a four-channel function generator, and an Arduino-based control unit for switching output channels and controlling motor rotation. **(B)** Schematic representation of control patterns achieved through various combinations of magnetic rotation and AC frequencies, enabling motion in four orthogonal directions. **(C)** Demonstration of rectangular trajectories achieved with 27 μm JPs using a manually applied control sequence. The experimental conditions included a 10 μM NaCl solution and 15 Vpp AC voltage, with the magnet rotating at 100 rpm (1.67 Hz). Scale bars: 100 μm. **(D)** Characterization of the velocity vector's horizontal (magnetic rolling) and vertical (electric propulsion) components, represented by the angle θ, as a function of the electric field frequency. Measurements were taken at 100 rpm rotation rate and 15 Vpp AC voltage. At low frequencies, ICEP dominates, producing primarily vertical motion. At intermediate frequencies, sDEP becomes significant, reversing the vertical motion direction while maintaining dominance. At high frequencies in the MHz range, electric propulsion diminishes, and horizontal magnetic rolling prevails.

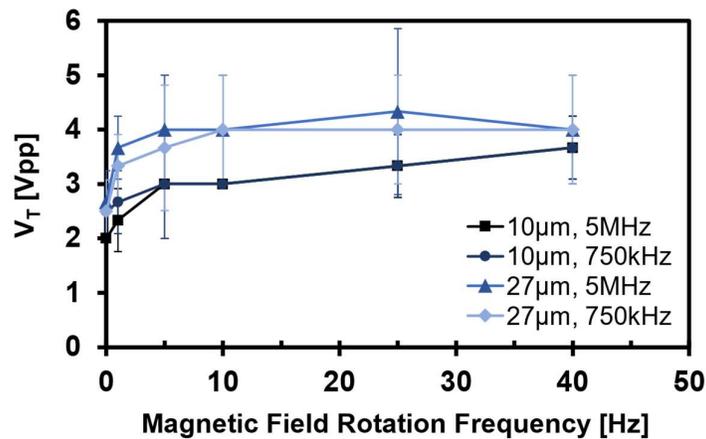

**Supplementary Fig. 6. Threshold voltage scan vs. magnetic field rotation rate.** Measurements of the threshold voltage ($V_T$) required to hold JPs at the chamber ceiling via electrostatic trapping were conducted as a function of the magnetic field rotation frequency. The scan was performed using 10 μm and 27 μm JPs at AC frequencies of 750 kHz and 5 MHz, where orthogonal electric propulsion is minimized. Compared to the idle state without magnetic rolling (0 Hz rotation), the threshold voltage generally increased by 1-2 Volts at lower rotation rates (<10 Hz) and remained relatively unchanged with further increases in rotation speed.

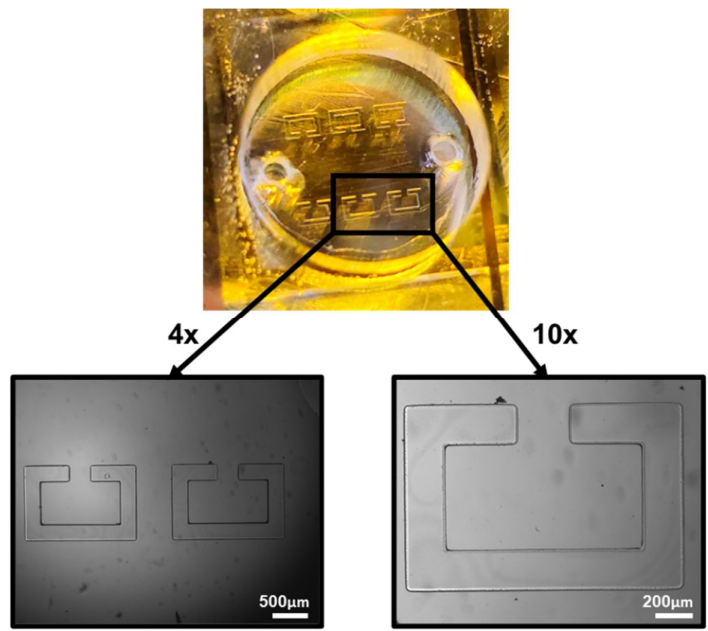

**Supplementary Fig. 7. Microchamber with SU-8 strcutures at 4x and 10x magnifications.** The U-shaped structures, approximately 50 μm in height, effectively block direct passage while incorporating an opening that permits the controlled microrobot to enter and exit the inner area through surface motion alone. The signicantly larger scale of the sturctures relative to the microrobot allows them to be reagrded as 3D obstacles upon closer inspection.

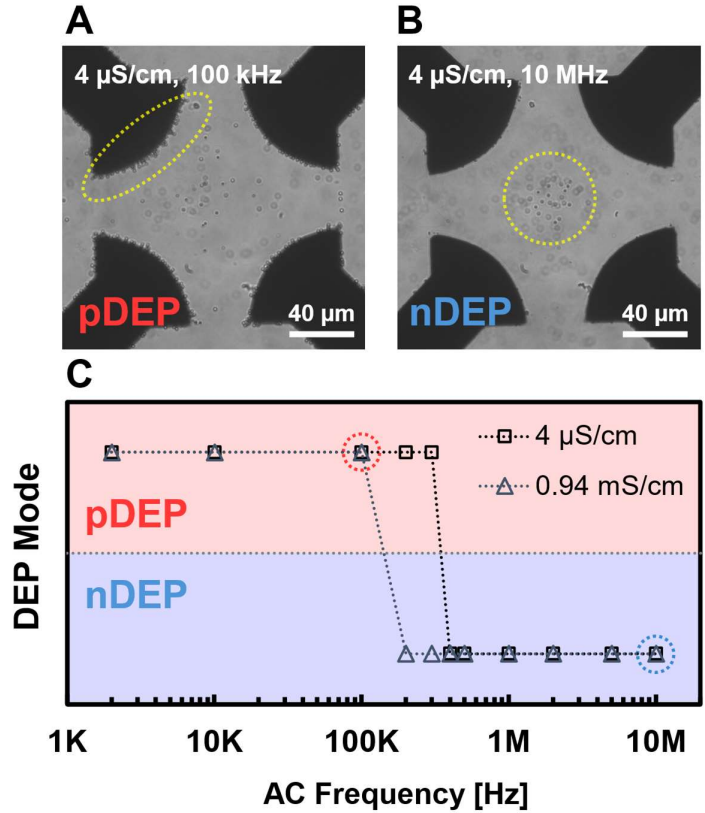

**Supplementary Fig. 8. DEP response of 1.1 μm diameter polystyrene (PS) particles within a quadrupolar electrode array. (A)** Image of the quadrupolar array at 40x magnificaiton showing the PS particles attracted to the electrodes (i.e. exhibiting pDEP response) at a solution conductivity of 4 μS cm$^{-1}$ and AC field of 100 kHz, 10 V$_{pp}$. **(B)** The same for an AC field of 10 MHz, 10 V$_{pp}$, with the PS particles pushed towards the center of the array (i.e. exhibit nDEP response). **(C)** Binary division of pDEP / nDEP response as a function of the applied 10 V$_{pp}$ AC frequency, at solution conductivities of 4 μS cm$^{-1}$ and 0.94 mS cm$^{-1}$. The cross over frequency occurs at ~350 kHz $\pm$ 50 kHz for 4 μS cm$^{-1}$, and ~150 kHz $\pm$ 50 kHz for 0.94 mS cm$^{-1}$.

**Supplementary Note 1. Analytical derivation of the threshold voltage for electrostatically holding the JP at the ceiling.**

Here we suggest a simple approach based on using the weak–field approximation, thin EDL and the linearized Poisson–Nernst–Planck (PNP) approach for computing the equivalent dipole of a metallo-dielectric Janus particle by imposing the proper boundary conditions on the two hemispheres[50,51]. On the metallic (perfect conductor) side[52] we employ the common RC boundary condition

$$\frac{\partial \phi}{\partial r} = -i\Omega_{RC}\phi, \qquad (1)$$

whereas on the dielectric side, we account for 'surface conductivity' and use the following[53],

$$\frac{\partial \phi}{\partial r} = -\widetilde{K}_s \nabla^2 \phi. \qquad (2)$$

where both (1) and 2 are enforced at $r = a$. Here $\Omega_{RC} = \frac{\omega \lambda_0 a}{D}$ represents the dimensionless angular frequency scaled by the corresponding RC frequency, $\widetilde{K}_s = \frac{K_s/a}{K_e - i\omega\epsilon_e} = \frac{(K_s/a)/K_e}{1 - i\Omega_{MW}}$ represents the dimensionless surface to bulk conductance, and $\Omega_{MW} = \frac{\omega \lambda_0^2}{D}$ represents the dimensionless angular frequency scaled by the corresponding Maxwell-Wagner frequency[54]. Herein, $\phi$ is the electric potential, $\lambda_0$ is the Debye length (EDL thickness), $\omega$ is the angular frequency of the applied field, $a$ is the radius of the JP, $D$ is the diffusion coefficient (symmetric electrolyte), $K_s$ is the surface conductance, $K_e$ is the solution bulk conductivity and $\epsilon_e$ is the permittivity of the medium.

Thus, the dipole of a homogenous metallic particle can be found from (1) as:

$$d_m = -\frac{1}{2} - \frac{3}{2}\frac{i\Omega_{RC}}{2 - i\Omega_{RC}}, \qquad (3)$$

and the dipole of a homogenous dielectric particle can be written in a similar way using (2) as:

$$d_e = -\frac{1}{2} + \frac{3}{2}\frac{\widetilde{K}}{\widetilde{K} - 1 + i\Omega_{MW}}, \qquad (4)$$

where $\widetilde{K} = (K_s/a)/K_e$.

The effective dipole of a metallo-dielectric JP can be approximated by the average between the distinct dipoles of the two hemispheres[51]. Thus, by including the effect of electrode screening with $\Omega_H = \frac{\omega \lambda_0 H}{D}$, with 2H representing the distance between the two planar electrodes, one gets[53]:

$$d_{eff} = \frac{1}{2}(d_m + d_e) \cdot \left[\frac{1}{1 + i/\Omega_H}\right] = \frac{\Omega_H}{4(i + \Omega_H)}\left[1 - \frac{6}{2 - i\Omega_{RC}} + \frac{3\widetilde{K}}{(\widetilde{K}-1) + i\Omega_{MW}}\right]. \qquad (5)$$

The frequency-dependent threshold voltage $V(\omega)$, for trapping the JP at the ceiling countering its own weight, can then be found by using the dipole-dipole approximation from

$$V(\omega) \sim \frac{1}{Re\{d_{eff}(\omega)\}} \sim 4\left(1 + \frac{1}{\Omega_H^2}\right)\left[1 - \frac{12}{4 + \Omega_{RC}^2} + \frac{3\widetilde{K}(\widetilde{K}-1)}{(\widetilde{K}-1)^2 + \Omega_{MW}^2}\right]^{-1}, \qquad (6)$$

since the electrostatic (attraction) force is proportional to $|V(\omega)d_{eff}(\omega)|^2$.

Next, recalling that, $\Omega_H \gg \Omega_{RC} \gg \Omega_{MW}$, then for small frequencies say $\Omega_H \sim O(1)$ eq.(6) yields

$$V(\omega) \sim \frac{4}{\Omega_H^2}, \qquad (7)$$

while for large frequencies $\Omega_{MW} \sim O(1)$, the contributions due to both $\Omega_H$ and $\Omega_{RC}$ (being large) can be neglected resulting in

$$V(\omega) \sim 4\left[1 + \frac{3\widetilde{K}(\widetilde{K}-1)}{(\widetilde{K}-1)^2 + \Omega_{MW}^2}\right]^{-1}. \tag{8}$$

Eq. (8) implies the existence of an asymptote $V(\omega) \sim 4$ for $\Omega_{MW} \gg 1$ and an inflection point (i.e. change of curvature) at $\Omega_{MW} = \sqrt{\frac{(\widetilde{K}-1)(4\widetilde{K}-1)}{3}}$.

For moderate frequencies, namely, $\Omega_{RC} \sim O(1)$, one gets $\Omega_H \gg 1$, $\Omega_{MW} \ll 1$ so that

$$V(\omega) \sim 4\left[1 + \frac{3\widetilde{K}}{(\widetilde{K}-1)} - \frac{12}{4+\Omega_{RC}^2}\right]^{-1}, \tag{9}$$

suggesting again that there is another inflection point at $\Omega_{RC} = 2\sqrt{\frac{\widetilde{K}+2}{3(4\widetilde{K}-1)}}$, in accordance with the experimental measurements.

**Supplementary Note 2. Theoretical calculations of the settling time for a descending JP dropped from the chamber ceiling.**

A spherical particle with a total mass $m_{tot}$, radius $R$ and volume $V_{tot}$, dropped at rest from an initial height $h$ inside a fluid with density $\rho_{fl}$ and viscosity $\mu$, will experience the following force balance on the vertical axis:

$$m_{tot}\frac{dv}{dt} = \Sigma F = m_{tot}g - F_b - F_d \tag{10}$$

Where $F_b = \rho_{fl}V_{tot}g$ is the buoyancy force and $F_d = 6\pi\mu Rv$ is Stokes' drag force.

A Janus particle (JP) made of spherical polystyrene (PS) core with density $\rho_{PS}$, coated by deposited metal layers of Cr, Ni, and Au on its top hemisphere, with densities $\rho_{Cr}, \rho_{Ni}, \rho_{Au}$ and layer thicknesses of $t_{Cr}, t_{Ni}, t_{Au}$, can be ideally modeled as a sphere with internal radius $r_{PS}$ capped by hemispherical shells with varying internal and external radii per metal layer, such that for layer $i$ the radii are

$$r_{in,i} = r_{out,i-1}, \qquad r_{out,i} = r_{in,i} + t_i \tag{11}$$

and for the first layer the radii are $r_{in} = r_{PS}, r_{out} = r_{PS} + t_{Cr}$.

For each hemispherical shell the internal volume is given by:

$$V_{shell} = \frac{2\pi}{3}(r_{out}^3 - r_{in}^3) \tag{12}$$

such that the mass of layer $i$ is $m_i = \rho_i V_{shell,i}$, and the total mass is:

$$m_{tot} = \rho_{PS} \frac{4\pi}{3} r_{PS}^3 + \sum_i m_i \tag{13}$$

The total volume of the JP can be calculated from the volume of the bottom PS hemisphere combined with the volume of the metallic hemisphere with a total radius of $r_{metal} = r_{PS} + \sum_i t_i$, such that:

$$V_{tot} = \frac{2\pi}{3}\left(r_{PS}^3 + r_{metal}^3\right) \tag{14}$$

and the effective density of the particle can be defined by $\rho_{eff} = \frac{m_{tot}}{V_{tot}}$.

Reorganizing eq. (10) and substituting $\rho_{eff}$ yields the equation:

$$\frac{dv}{dt} + \frac{6\pi\mu R}{m_{tot}} v = g\left(1 - \frac{\rho_{fl}}{\rho_{eff}}\right), \quad v(t=0) = 0 \tag{15}$$

which is solved for the velocity by:

$$v(t) = \frac{m_{tot} g \left(1 - \frac{\rho_{fl}}{\rho_{eff}}\right)}{6\pi\mu R}\left(1 - e^{-\frac{6\pi\mu R}{m_{tot}}t}\right) = v_t \left(1 - e^{-\frac{t}{\tau}}\right) \tag{16}$$

where $v_t$ is the terminal velocity and $\tau$ is the characteristic time.

Recalling that $h = H - 2R$, where $H$ is the height of the microchamber, for small enough $\tau$ the exponential term can be neglected, yielding the settling time:

$$t_{settling} = \frac{H - 2R}{v_t} \tag{17}$$

Substituting relevant densities, layer thicknesses and fluid (water) properties from **Supplementary Table 2**, we can estimate the terminal velocities and settling times of JPs and bare PS particles with diameters of 10 μm and 27 μm. The effective radius R for Stokes' drag and eq. (17) is taken as the slightly larger radius of the metal hemisphere for the JPs, i.e. $R = r_{metal}$, and the nominal sphere radius $R = r_{PS}$ for bare PS particles. The results, presented in **Supplementary Table 1**, show that the characteristic time for reaching terminal velocity is indeed negligible compared to the settling time.

**Supplementary Table 1. Theoretical analysis results for JP and bare PS particles settling time.**

| Particle | Characteristic Time, $\tau$ | Terminal Velocity, $v_t$ | Settling Time, $t_{settling}$ |
| --- | --- | --- | --- |
| 10 µm JP | 7.22 µs | 15.83 µm/s | 6.93 s |
| 10 µm PS | 5.83 µs | 2.72 µm/s | 40.36 s |
| 27 µm JP | 46.25 µs | 55.19 µm/s | 1.68 s |
| 27 µm PS | 42.52 µs | 19.86 µm/s | 4.68 s |

**Supplementary Table 2. Parameters for the theoretical calculations.**

| Parameter | Value |
| --- | --- |
| $g$, gravitational constant | 9.81 [m/s²] |
| $\rho_{fl}$, Density of water | 1000 [kg/m³] |
| $\rho_{PS}$, Density of polystyrene | 1050 [kg/m³] |
| $\rho_{Cr}$, Density of chromium | 7192 [kg/m³] |
| $\rho_{Ni}$, Density of nickel | 8907 [kg/m³] |
| $\rho_{Au}$, Density of gold | 19283 [kg/m³] |
| $\mu$, Viscosity of water | $10^{-3}$ [Pa*s] |
| $t_{Cr}$, Thickness of the chromium layer | 15 [nm] |
| $t_{Ni}$, Thickness of the nickel layer | 55 [nm] |
| $t_{Au}$, Thickness of the gold layer | 15 [nm] |
| $H$, Height of the microchamber ceiling | 120 [µm] |
| $K_e$, Bulk solution conductivity | 4 [µS/cm] |
| $\epsilon_0$, Electric permittivity of vacuum | $8.854 * 10^{-12}$ [C*V/m] |
| $\epsilon$, Solution electric permittivity | $80 * \epsilon_0$ [C*V/m] |
| $D$, Diffusion coefficient | $2*10^{-9}$ [m²/s] |
| $\lambda_0$, Debye length | $\sqrt{\frac{\epsilon D}{K_e}} = 59.5$ [nm] |